\documentclass[lettersize,journal]{IEEEtran}
\usepackage{subcaption}
\usepackage{graphicx}

\usepackage{pifont}
\newcommand{\cmark}{\color{green}\ding{52}}%
\newcommand{\xmark}{\color{red}\ding{55}}%

\usepackage{amsmath,amsfonts}
\usepackage{algorithmic}
\usepackage{algorithm}
\usepackage{array}
\usepackage{textcomp}
\usepackage{stfloats}
\usepackage{url}
\usepackage{verbatim}
\usepackage{cite}
\usepackage{mathtools,lipsum}
\usepackage{cuted}
\usepackage{footnote}
\makesavenoteenv{tabular}
\usepackage{booktabs}
\usepackage{setspace}
\usepackage{tablefootnote}
\usepackage{caption}
\usepackage{multirow}
\usepackage{wrapfig}
\usepackage{multicol}
\usepackage{tikz}
\usepackage{pdfpages}

\def\vec#1{\mathchoice{\mbox{\boldmath$\displaystyle#1$}}
{\mbox{\boldmath$\textstyle#1$}}
{\mbox{\boldmath$\scriptstyle#1$}}
{\mbox{\boldmath$\scriptscriptstyle#1$}}}

\begin{document}

\title{From Hand-Crafted Metrics to Evolved Training-Free Performance Predictors for Neural Architecture Search via Genetic Programming}

\author{\IEEEauthorblockN{
Quan Minh Phan,
Ngoc Hoang Luong
}\\
\IEEEauthorblockA{University of Information Technology, Ho Chi Minh City, Vietnam}\\
\IEEEauthorblockA{Vietnam National University, Ho Chi Minh City, Vietnam}\\
\{quanphm, hoangln\}@uit.edu.vn
}

\markboth{Journal of \LaTeX\ Class Files,~Vol.~14, No.~8, August~2021}%
{Shell \MakeLowercase{\textit{et al.}}: A Sample Article Using IEEEtran.cls for IEEE Journals}

\IEEEpubid{0000--0000/00\$00.00~\copyright~2021 IEEE}

\maketitle

\begin{abstract}
Estimating the network performance using zero-cost (ZC) metrics has proven both its efficiency and efficacy in Neural Architecture Search (NAS).
However, a notable limitation of most ZC proxies is their inconsistency, as reflected by the substantial variation in their performance across different problems.
Furthermore, the design of existing ZC metrics is manual, involving a time-consuming trial-and-error process that requires substantial domain expertise.
These challenges raise two critical questions: \textit{(1) Can we automate the design of ZC metrics?} and \textit{(2) Can we utilize the existing hand-crafted ZC metrics to synthesize a more generalizable one?}
In this study, we propose a framework based on Symbolic Regression via Genetic Programming to automate the design of ZC metrics.
Our framework is not only highly extensible but also capable of quickly producing a ZC metric with a strong positive rank correlation to true network performance across diverse NAS search spaces and tasks.
Extensive experiments on 13 problems from NAS-Bench-Suite-Zero demonstrate that our automatically generated proxies consistently outperform hand-crafted alternatives.
Using our evolved proxy metric as the search objective in an evolutionary algorithm, we could identify network architectures with competitive performance within 15 minutes using a single consumer GPU. 
\end{abstract}


\section{Introduction}

\begin{figure}[htbp]
    \centering
    \includegraphics[width=\linewidth]{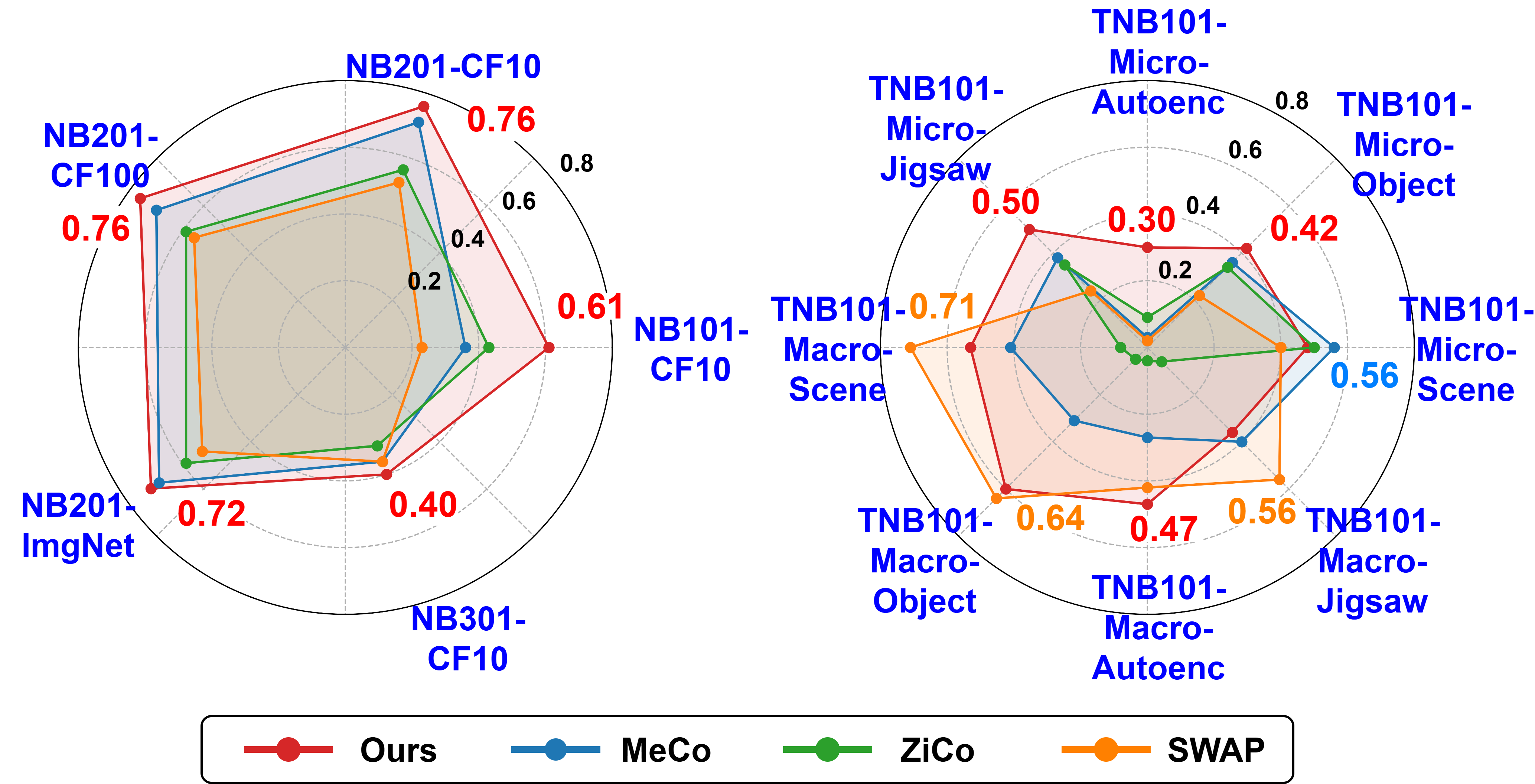}
    \caption{Comparisons in Kendall's Tau rank correlations between our automatically-designed ZC metric and state-of-the-art ZC metrics (i.e., MeCo, ZiCo, and SWAP) on NAS-Bench-101/201/301 (Left) and TransNAS-Bench-101-Micro/Macro (Right).
    Highest Kendall's $\tau$ scores for each problem are presented.
    The results highlight the consistency of our ZC metric across various search spaces and tasks.
    }\label{fig:thumbnail}
\end{figure}
Neural architecture search (NAS), a subfield of Automated Machine Learning (AutoML), focuses on the automatic design of high-performance network architectures~\cite{nas_survey}.
NAS can be described as a procedure in which a \emph{search algorithm} explores a \emph{search space} containing potential architectures.
During the exploration, the search algorithm assesses the quality of candidate networks using a \emph{performance estimation strategy}.
Early NAS studies estimated the quality of networks based on their performance on the validation dataset~\cite{nas_as_rl, rea}.
Although top-performing networks could be obtained, a single NAS run took several weeks or months to complete and heavily depended on hardware resource (i.e., 800 GPUs for a single run in~\cite{rea}).
Recent works have proposed various approaches to deal with this bottleneck such as weight-sharing~\cite{weight_sharing}, performance prediction~\cite{nas_predictor}, or zero-cost metrics~\cite{zc_nas}.
Among these techniques, using \emph{zero-cost} metrics (also known as \emph{training-free} metrics in the literature) offers the most efficiency since it can shorten the search time from days to a dozen seconds or minutes.
For example, \cite{swap} found a network architecture with competitive performance on CIFAR-10 within 6 minutes using SWAP, which is a zero-cost proxy that measures the expressivity of the network.
However, a significant limitation of the zero-cost metrics (ZC metrics) is their \emph{order-preserving ability}.
The order-preserving ability of ZC metrics can be assessed by examining the consistency between the rankings of architectures produced by ZC proxies and those based on their true performance~\cite{Zhang_2024_CVPR}.
Many studies have highlighted inconsistencies in network rankings based on ZC metrics' scores across different problems~\cite{zc_nas, nb_suite_zero}.
While ZC metrics exhibit positive rank correlations with network performance on some problems, they may demonstrate negative correlations on others.
Recent studies have attempted to address this issue when newly proposed ZC proxies (e.g., ZiCo~\cite{zico}, MeCo~\cite{meco}, SWAP~\cite{swap}) achieve positive rank correlations across all testing problems.
In these cases, a network that receives a high ZC score is also likely to exhibit high performance, and \emph{vice versa}.
However, despite achieving positive rank correlations, the scores for some problems remain close to 0.0, indicating a lack of correlation between the ZC scores and the networks' performance.
In this study, our target is to design a ZC metric that yields not only \textbf{the consistent order-preserving ability} but also \textbf{a high positive rank correlation} across various problems\footnote{In this paper, we define an NAS problem consisting of a single search space (e.g., NAS-Bench-101, DARTS) and a single task (e.g., image classification on CIFAR-10 dataset, image classification on ImageNet dataset, object detection on Taskonomy dataset). If we search network architectures within one search space but evaluate them on two different tasks, we consider that we are solving two NAS problems. We list all 13 NAS problems in our experiments in Appendix D of the Supplementary.}.

\begin{table*}[tbh]
\caption{List of frameworks that automatically design ZC metrics.}
\label{tab:difference}
\setstretch{1.15}
\centering
{
\begin{tabular}{lcccc}
\toprule
Name & Method & Features & \begin{tabular}[c]{@{}c@{}}Multiple-\\problems\end{tabular} & \begin{tabular}[c]{@{}c@{}}High-\\extensibility\end{tabular} \\ \midrule
EZNAS~\cite{ez_nas} & Symbolic Regression & Low-level & \xmark & \xmark \\
Auto-Prox~\cite{autoprox} & Symbolic Regression & Low-level & \cmark & \xmark \\
AZNAS~\cite{az_nas} & - & High-level & \xmark & \cmark \\
UP-NAS~\cite{up_nas} & Neural network & High-level & \xmark & \cmark \\\
RoBoT~\cite{robot} & Bayesian Optimization & High-level & \xmark & \cmark \\
\midrule
\textbf{Ours} & Symbolic Regression & High-level & \cmark & \cmark \\
 \bottomrule
\end{tabular}
}
\end{table*}
Most proposed ZC metrics are manually designed with expert knowledge.
\emph{Can we automate the design of efficient ZC metrics to form the proxies that are better than hand-crafted ones?}
The practicability of this approach has been demonstrated in several studies.
EZNAS~\cite{ez_nas} implemented Symbolic Regression to form a ZC metric by combining three low-level features of networks: weights, activations, and gradients.
While EZNAS only considered convolutional neural networks, Auto-Prox~\cite{autoprox} extended the framework to design ZC metrics for vision transformer architectures.
Additionally, a single run of Auto-Prox aims to search for a ZC metric that could perform well across multiple problems instead of a single problem as in EZNAS.
UP-NAS~\cite{up_nas} designed a new ZC metric by combining higher-level features of networks (i.e., hand-crafted ZC metrics) and taking their weighted sum, in which the weights of metrics are found by using a neural network.
However, a search process is needed to obtain suitable weights whenever we handle a new problem.
In this paper, we focus on the automatic designing of efficient ZC metrics.
Table~\ref{tab:difference} highlights the differences between our proposed framework and previous approaches that also automate the design of ZC metrics.
We discuss in detail the differences in Section~\ref{sec:proposed}.
Generally, our framework is designed to synthesize a new ZC metric from existing hand-crafted ones, and the outcome metric could perform well across diverse problems.
Our method offers high extensibility, making it straightforward to adapt to various problems or modify input features to enhance the performance of the resulting metric.
Our \textbf{main contributions} are as follows:
\begin{itemize}
    \item We propose a highly extensible framework for automatically designing a ZC metric that not only consistently preserves the order of network performance but also achieves a high positive rank correlation across multiple problems.
    \item Our framework can synthesize a new ZC metric within only 10 minutes, which is significantly faster than the 24 hours required by EZNAS.
    Compared to previous ZC metrics, including newly proposed ones (i.e., ZiCo, MeCo, SWAP), our metric achieves the state-of-the-art Kendall's tau correlation for NAS-Bench-101/201/301 search spaces and comparable scores for TransNAS-Bench-101-Micro/Macro search spaces (see Fig.~\ref{fig:thumbnail}).
    \item The performance of the best networks found by our synthesized ZC proxy surpasses those found by hand-crafted ZC proxies in 8 out of 13 problems in NAS-Bench-Suite-Zero.
    By integrating the found ZC metric with the Aging Evolution~\cite{rea} algorithm, we could find a network with comparable performance on the CIFAR-10 dataset within 15 minutes, demonstrating the practical applicability of our obtained ZC metric in real-world NAS scenarios.
\end{itemize}

\section{Related Work}
\subsection{Limitation of ZC-NAS metrics}
Zero-cost metrics are widely employed in NAS due to their ability to estimate the network performance with a trivial cost~\cite{nb_suite_zero}.
However, several studies have indicated their inconsistent order-preserving ability across different problems~\cite{zc_nas, nb_suite_zero}.
For example, the ZC metric \textbf{Grad-norm} has the Spearman rank correlation to network performance of $0.58$ for the NAS-Bench-201 search space but has a score of $-0.21$ for the NAS-Bench-NLP search space~\cite{zc_nas}.
Recent proposed ZC metrics (e.g., \textbf{ZiCo}, \textbf{MeCo}, \textbf{SWAP}) have tackled this issue since they consistently yield a positive rank correlation across various problems.
Nevertheless, their rank correlation scores are close to $0.0$ for some cases (e.g., the \textbf{ZiCo} metric for the TransNAS-Bench-101-Macro search space), exhibiting the lack of correlation between the ZC metrics' scores and the networks' performance.
Designing a ZC metric that exhibits a high correlation to networks' performance and a strong order-preserving ability across various problems is essential.

\subsection{Automatic designing of ZC-NAS metrics}
While most ZC metrics are manually designed with expert knowledge, several studies aim to design ZC metrics \emph{automatically}~\cite{ez_nas, autoprox, up_nas, robot, az_nas}.
\textbf{EZNAS}~\cite{ez_nas} proposed a framework that uses Symbolic Regression (SR) to synthesize a ZC metric from three basic statistics of convolutional neural networks: weights, activations, and gradients.
The empirical results shown that the ZC metric found by {EZNAS} could surpass hand-crafted metrics.
\textbf{Auto-Prox}~\cite{autoprox} then modified the framework of {EZNAS} to find a ZC metric for vision transformer networks.
In {EZNAS}, the dataset used by SR consists of networks' information in a single problem.
Such an approach might make the resulting ZC metric overfitted on the experimented problem and perform worse on the unseen problems.
{Auto-Prox} solved this issue by combining multiple problems into the dataset of SR instead of a single one.
However, {Auto-Prox} only tested the proposed framework in finding the ZC proxy for candidate networks within the same search space.
When facing a different search space, the search process was re-conducted to synthesize a new ZC proxy.
On the other hand, both {EZNAS} and {Auto-Prox} are restricted from discovering more robust ZC metrics when they synthesize new ZC proxies using \emph{low-level} features of the network.
Compared to {EZNAS} and {Auto-Prox}, the extensibility of \textbf{AZ-NAS}~\cite{az_nas}, \textbf{UP-NAS}~\cite{up_nas}, and \textbf{RoBot}~\cite{robot} are higher when they synthesized new ZC metrics from \emph{high-level} features: hand-crafted ZC metrics.
\textbf{AZ-NAS}~\cite{az_nas} aims to compute a weighted sum of four different ZC metrics representing the expressivity, progressivity, trainability, and complexity of networks, where the weights are dynamically optimized during each NAS run for solving a specific NAS problem.
\textbf{UP-NAS}~\cite{up_nas} assumes there is a linear relationship between hand-crafted ZC metrics and uses a neural network to find the weights of 13 ZC metrics listed in NAS-Bench-Suite-Zero.
A new ZC proxy was then formed by taking their weighted sum.
The approach of \textbf{RoBoT}~\cite{robot} is similar to UP-NAS but the weights of ZC metrics are obtained through Bayesian Optimization.
However, the limitation of above methods is similar to {EZNAS} as theirs dataset only covers a single problem.
A new ZC metric thus needs to be searched for whenever handling a different NAS problem.

\subsection{NAS Benchmarks}
An NAS benchmark can be viewed as a database consisting of essential information about candidate network architectures (e.g., train/validation/test performance, the number of parameters) within the same search space.
The \textbf{NAS-Bench-101/201/301} benchmarks~\cite{nb101, nb201, nb301} provide the performance of candidate networks for the image classification task with different datasets.
The \textbf{TransNAS-Bench-101-Micro/Macro} benchmarks~\cite{tnb101} consider the network performance on other tasks beyond the image classification such as scene classification, and object detection.
The \textbf{NAS-Bench-Suite-Zero} benchmark~\cite{nb_suite_zero} is the collection of the aforementioned benchmarks and additionally computes the scores for all candidate networks in terms of 13 various ZC metrics.
The purpose of using NAS benchmarks is that we can quickly evaluate the effectiveness of different NAS methods and fairly compare them under the same setting.
In this paper, besides using NAS benchmarks to assess our proposed method, we utilize NAS benchmarks to create the dataset for our framework to automatically learn new ZC metrics.

\section{Proposed Framework}\label{sec:proposed}
Our proposed framework for automating the design of ZC-NAS metrics is based on Symbolic Regression (SR).
Given existing hand-crafted ZC metrics, we perform SR to search for the most effective way to combine these metrics using mathematical operators (e.g., \emph{add}, \emph{mul}).
In this section, we first outline our approach to building the dataset for putting into SR in Section~\ref{ssec:training_data}.
The proposed mechanism for evaluating the quality of ZC metrics during the search process is then detailed in Section~\ref{ssec:fitness_eval}.
The method for representing the synthesized ZC proxies and the ways that SR discovers novel ZC proxies are presented in Sections~\ref{ssec:representation} and \ref{ssec:variation_ops}, respectively.
The entire search procedure of our framework is discussed in Section~\ref{ssec:procedure}.
While the use of SR for automatically designing ZC metrics was introduced in EZNAS~\cite{ez_nas} and Auto-Prox~\cite{autoprox}, there are notable differences between our framework and theirs, particularly in the construction of the input dataset for SR, the mechanism for evaluating synthesized ZC metrics, and the search procedure.
These differences are thoroughly discussed in the following sections.

\subsection{Dataset Building}\label{ssec:training_data}
Our approach to building the input dataset for SR closely resembles that of UP-NAS~\cite{up_nas}, which uses hand-crafted ZC-NAS metrics as feature variables and validation performance of networks as the ground truth.
However, while the dataset of UP-NAS only covers a single problem, our dataset encompasses the networks' information across \emph{multiple problems}.
Although the suggestion of integrating multiple problems into the input dataset of SR was proposed in Auto-Prox~\cite{autoprox}, Auto-Prox uses the low-level features (i.e., network statistics), whereas we utilize the high-level features (i.e., hand-crafted ZC metrics).

When creating the input dataset, while the cost of computing the ZC proxy values is negligible, the most significant computational cost arises from defining the ground truth (i.e., the network performance).
However, we overcome this obstacle by employing existing NAS benchmarks (e.g., NAS-Bench-101, NAS-Bench-201), which provide the performance of numerous candidate architectures across various search spaces and tasks.
Using NAS benchmarks also demonstrates strong extensibility.
Whenever a new NAS benchmark or hand-crafted ZC metric is introduced, we can quickly compute the ZC metric scores of networks at a negligible cost and enrich the input dataset by integrating the new information.
Another way of defining the ground truth involves using supernets (e.g., Once-for-All~\cite{ofa}) or performance predictors (e.g., XGBoost~\cite{nas_predictor}), for which configurations and weights are available.
In this study, we primarily focus on NAS benchmarks to utilize truly trained performance rather than predicted performance.

\subsection{Objective Evaluation}\label{ssec:fitness_eval}
Our target is to design a ZC metric that ranks candidate networks in the same order as their true performance.
We thus use Kendall's $\tau$ rank correlation to evaluate the quality of ZC metrics generated by SR during the search process, similarly to EZNAS, UP-NAS, and Auto-Prox.
However, the evaluation methods used in EZNAS and UP-NAS are not applicable to our framework, as our dataset comprises multiple problems rather than a single one.
Auto-Prox, which also uses multiple problems to build the input dataset, evaluates the quality of an arbitrary ZC metric by computing its Kendall's $\tau$ value for each problem and taking the weighted sum.
This approach has its limitation.
In particular, because the weight for each problem must be \emph{pre-defined} before searching, it may cause the resulting ZC metric to be biased towards the problem with the highest weight.
Setting equal weights for all problems is not effective due to the differences in Kendall's $\tau$ values across problems, leading SR to focus disproportionately on the problem that produces the highest rank correlation and neglect others.

Our proposed mechanism for evaluating ZC metrics generated by SR is as follows.
Given $\{\tau_1, \tau_2, \ldots, \tau_N\}$ as the set of Kendall's $\tau$ values of an arbitrary ZC metric $\vec{x}$ on $N$ problems, the quality of $\vec{x}$ (denoted as $Score(\vec{x})$) is defined as:
\begin{equation}\label{eqn:score}
    Score(\vec{x}) = \sum_{i = 1}^{N} \frac{\tau_i - \tau^-_i}{\tau^+_i - \tau^-_i},
\end{equation}
where $\tau^-_i$ and $\tau^+_i$ represent the lowest and highest Kendall's $\tau$ values obtained so far for the $i$-th problem.
Equation \ref{eqn:score} scales Kendall's $\tau$ score for each problem into the range $[0, 1]$, where the new value being $0$ and $1$ if it corresponds to the lowest and highest scores obtained so far for the problem, respectively.
The maximum score for the ZC metric $\vec{x}$ is therefore $N$, if it yields the highest Kendall's $\tau$ values for all problems at the time of evaluation.

\subsection{Solution Representation}\label{ssec:representation}
\begin{figure}[!htbp]
    \centering
    \includegraphics[width=0.8\linewidth]{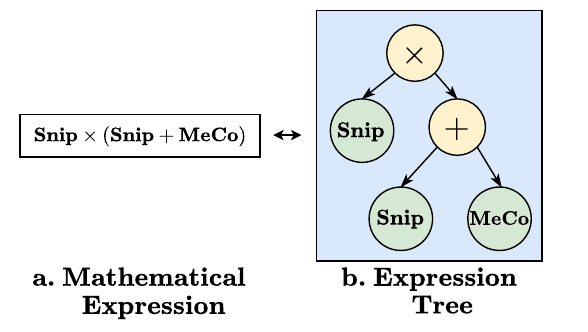}
    \caption{Example of using expression tree to represent the expression $\textbf{Snip} \times (\textbf{Snip} + {\textbf{MeCo}})$ in Symbolic Regression.}\label{fig:ET_example}
\end{figure}
Our use of SR to design the ZC metric is similar to finding the optimal function that combines multiple hand-crafted ZC metrics, which can be represented as an expression tree.
Fig.~\ref{fig:ET_example} illustrates an example of a function combining two ZC proxies and the corresponding expression tree.
In an expression tree, the \emph{leaf} nodes represent variables or constants (e.g., nodes `\textbf{Snip}' and `\textbf{MeCo}' in Fig.~\ref{fig:ET_example}), while the \emph{internal} nodes represents mathematical operators (e.g., nodes `$+$' and `$\times$' in Fig.~\ref{fig:ET_example}).
In our experiments, we constrain the search space by using a fixed set of primitive mathematical operators $\{$\emph{add}, \emph{sub}, \emph{mul}, \emph{div}, \emph{neg}, \emph{log}, \emph{sqrt}$\}$ (more details on these operators are provided in Appendix C of the Supplementary), and we exclude constants in the mathematical expressions, meaning that the leaf nodes are exclusively ZC metrics.
The minimum and maximum tree depths are also restricted to 2 and 10, respectively.

\subsection{Variation Operators}\label{ssec:variation_ops}
\begin{figure}[!htbp]
    \centering
    \includegraphics[width=\linewidth]{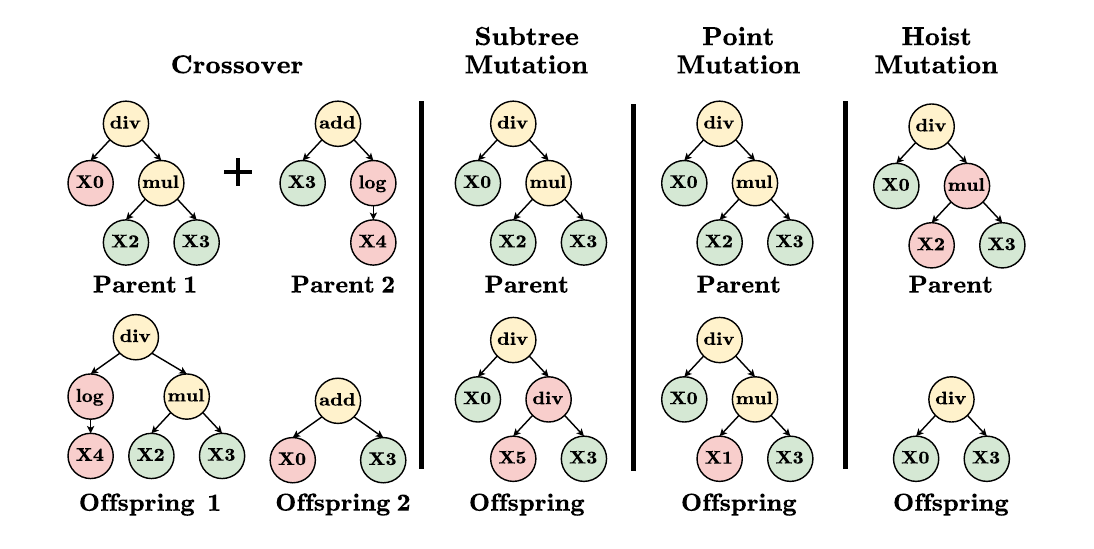}
    \caption{Illustration of crossover and mutation operators for expression trees.}
    \label{fig:variation_operators}
\end{figure}
We implement basic variation operators, including crossover, subtree mutation, hoist mutation, and point mutation, to enable genetic programming to discover novel expression trees (i.e., ZC metrics) during the search process.
Illustrations of these operators are exhibited in Fig.~\ref{fig:variation_operators}.

\subsection{Search Procedure}\label{ssec:procedure}
\begin{figure*}[!ht]
    \centering
    \includegraphics[width=\textwidth]{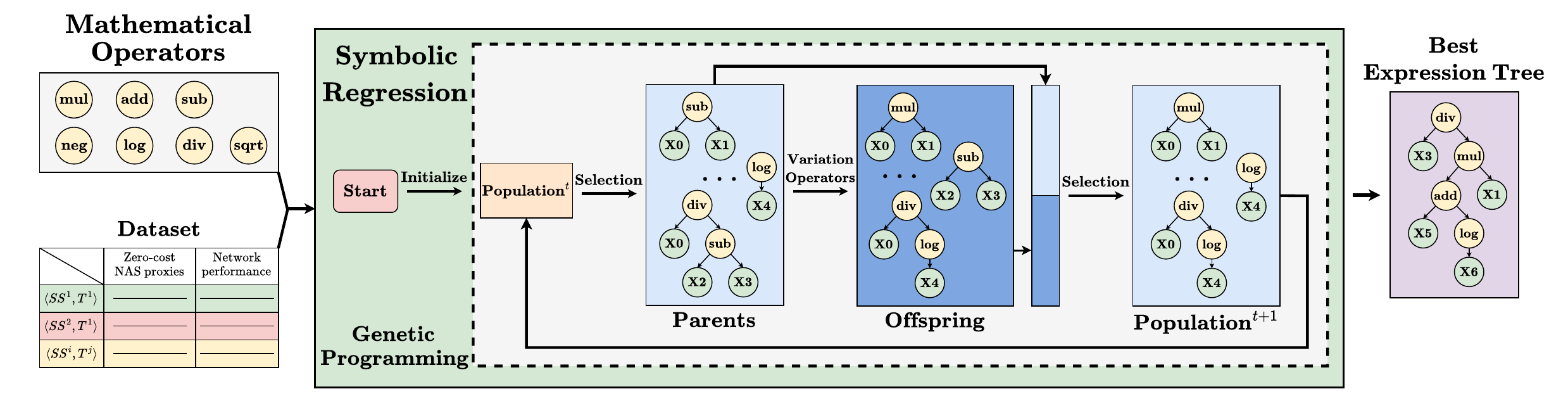}
    \caption{The procedure of searching new ZC metrics with Symbolic Regression via Genetic Programming. {Our dataset consists of multiple NAS problems and it thus contains multiple search spaces $SS$ = \{$SS_1$, $SS_2$, $\ldots$, $SS_i$\} and tasks $T$ = \{$T_1$, $T_2$, $\ldots$, $T_i$\}. $SS_i$ refers to the $i$-th search space in the list of search spaces $SS$ and $T_j$ refers to the $j$-th in the list of tasks $T$.}}
    \label{fig:procedure}
\end{figure*}
Fig.~\ref{fig:procedure} illustrates the entire procedure we use SR for synthesizing new ZC metrics in this study.
First, we initialize a population of $N$ synthesized ZC metrics, which are represented by expression trees as described in Section \ref{ssec:representation}.
We then evaluate the quality of these ZC metrics using the method detailed in Section \ref{ssec:fitness_eval} and employ binary tournament selection to choose parents for generating offspring expressions.
Given a set $S$ of $N$ individuals, we divide $S$ into $N/2$ pairs and select the winner from each pair based on their quality.
This process is repeated until $N$ parent expressions (i.e., winners) are selected.
The parent expressions then undergo the variation operators (i.e., crossover and mutation, as discussed in Section \ref{ssec:variation_ops}) to produce offspring expressions.
The offspring are evaluated for their quality, and along with the parents, form a set $P+O$ with a size of $2\times N$.
We again perform tournament selection on the set $P+O$ to select the $N$ best expression trees for the population in the next generation.
The search process of SR is executed until the stopping criterion is satisfied (e.g., reaching the maximum number of generations), and the final output of SR is the best expression tree in the final population.

We highlight the differences between our search procedure and those used in EZNAS and Auto-Prox.
In their frameworks, after the offspring are produced and evaluated for quality, they are immediately chosen as individuals for the next population, without comparing to the parents as in our approach.
The drawback of such non-elitist approaches is that good solutions could be inadvertently eliminated and replaced with inferior ones.
In contrast,
by comparing parents and offspring, we ensure that high-quality ZC metrics remain in the population until the end of the search process.

\section{Experiments and Results}
The majority of our experiments are conducted on \textbf{NAS-Bench-Suite-Zero}~\cite{nb_suite_zero}, which consists of five different NAS benchmarks: \textbf{NAS-Bench-101}~\cite{nb101}, \textbf{NAS-Bench-201}~\cite{nb201}, \textbf{NAS-Bench-301}~\cite{nb301}, and \textbf{TransNAS-Bench-101-Micro/Macro}~\cite{tnb101}).
The network architectures in \textbf{NAS-Bench-101/201/301} are evaluated for the \textbf{Image classification} task, while those in \textbf{TransNAS-Bench-101-Micro/Macro} are evaluated on the tasks of \textbf{Object detection}, \textbf{Scene classification}, \textbf{Jigsaw puzzle}, and \textbf{Autoencoding}\footnote{There are three additional tasks \textbf{Room Layout}, \textbf{Surface Normal} and \textbf{Semantic Segmentation} for the two TransNAS-Bench-101 search spaces, but we encountered a dataset issue similar to \cite{swap}.}.
In summary, we experiment and validate our proposed method on \textbf{13} problems in NAS-Bench-Suite-Zero (more details of each problem are provided in Appendix D of the Supplementary).
{Beyond NAS benchmarks, we also demonstrate the effectiveness of our SR-designed ZC metric in practice by testing it on large-scale search spaces (i.e., DARTS~\cite{darts}, Once-For-All~\cite{ofa}) and dataset (i.e., ImageNet)}.
All experiments are conducted on a single GeForce RTX 3090 GPU.



\subsection{Searching for a robust ZC metric across multiple problems}\label{ssec:result_1}
We first run the proposed framework with the input dataset containing the values of 16 hand-crafted ZC-NAS proxies and the validation accuracies of network architectures across \emph{three} search spaces: NAS-Bench-101, NAS-Bench-201, and NAS-Bench-301.
Both the ZC scores and validation accuracies are measured on the CIFAR-10 dataset.
For ZC proxies, we reuse the values of 13 ZC metrics that are logged in NAS-Bench-Suite-Zero, including \textbf{FLOPs}, \textbf{Params}, \textbf{Jacov}~\cite{jacov}, \textbf{NWOT}~\cite{nwot}, \textbf{Synflow}~\cite{synflow}, \textbf{Snip}~\cite{snip}, \textbf{EPE-NAS}~\cite{epe_nas}, \textbf{Fisher}~\cite{fisher}, \textbf{Grad-norm}~\cite{zc_nas}, \textbf{Grasp}~\cite{grasp}, \textbf{L2-norm}~\cite{zc_nas}, \textbf{Zen}~\cite{zen}, \textbf{Plain}~\cite{zc_nas}).
Additionally, we compute the values of three state-of-the-art ZC metrics by using their published source code: \textbf{ZiCo}~\cite{zico}, \textbf{MeCo}~\cite{meco}, and \textbf{SWAP}~\cite{swap}.
For each search space, we add 70\% of the total networks into the input dataset for SR and use the remaining 30\% for the test dataset.
We run and test our framework for 31 independent runs with different seeds.
The population size of SR is set to 100 and the search process is terminated when SR reaches 50 generations.
Consequently, we explore a total of 5,000 synthesized ZC metrics at each run.
Additional hyperparameters for variation operators (e.g., the crossover and mutation probabilities) are presented in Appendix E of the Supplementary.

The obtained results reveal differences in the formulas of the ZC metrics designed by our framework across 31 independent runs (see Appendix I of the Supplementary).
However, the quality of these expressions is relatively consistent, as indicated by the small standard deviations (i.e., approximately 0.01 to 0.03, see Appendix A of the Supplementary) over 31 runs.
The results also show the similarity in the rank correlation of SR-designed ZC metrics on the input and test datasets, suggesting that the designed metrics are not overfitted to the input data and perform well across the entire dataset.
We also observe the differences in Kendall's $\tau$ scores achieved across different problems.
While we can obtain an average score of 0.74 for the NB201-CF10 problem, the scores for NB101-CF10 and NB301-CF10 problems are approximately 0.56 and 0.38, respectively.
Nonetheless, it is noteworthy that our $\tau$ values are significantly higher than those achieved by hand-crafted ZC proxies for these problems (comparisons are provided in Section \ref{ssec:result_2}).
\begin{figure}[htbp]
    \centering
    \includegraphics[width=\linewidth]{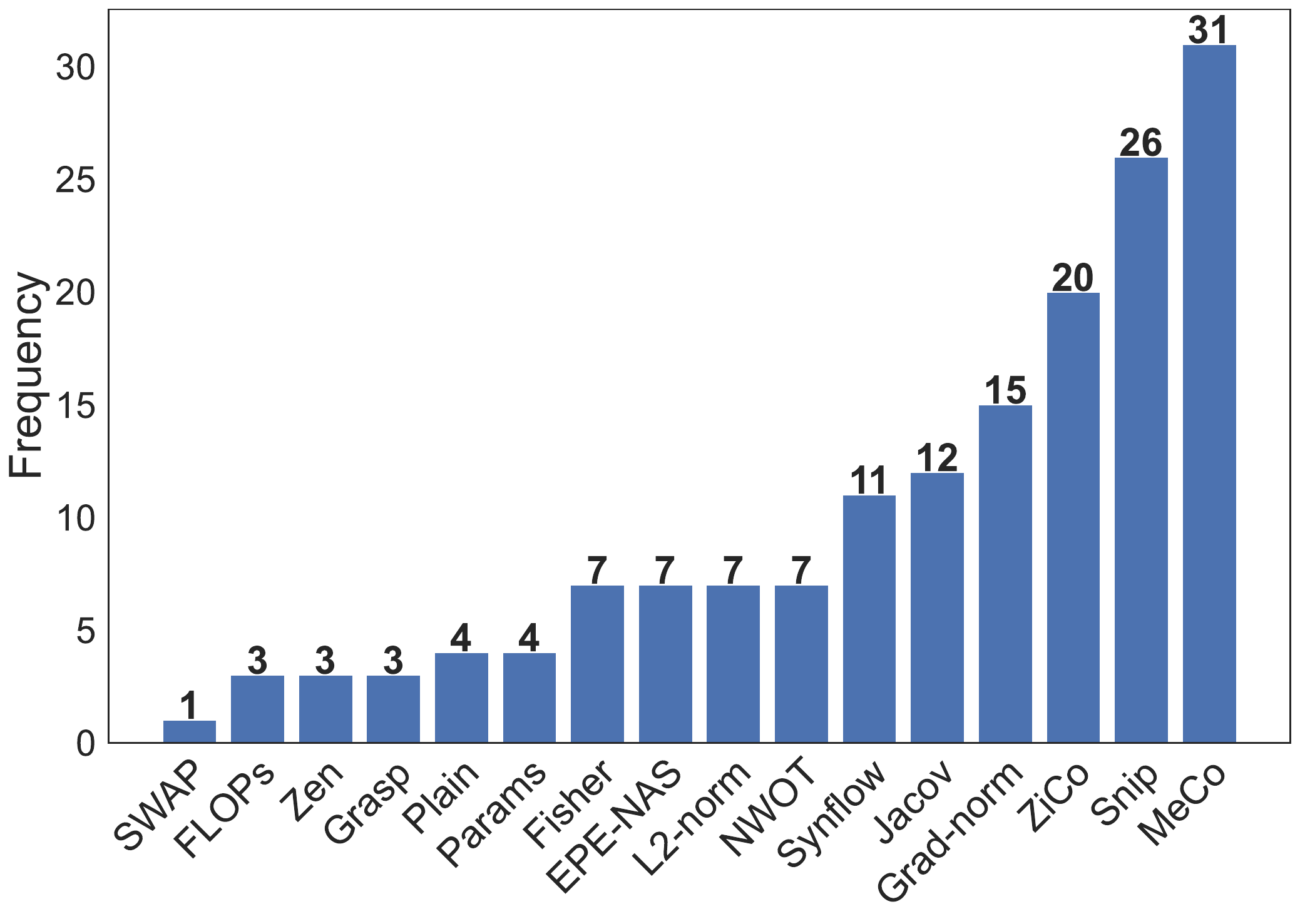}
    \caption{Frequency of hand-crafted ZC metrics in the 31 final ZC metrics synthesized by SR.}\label{fig:freq_1}
\end{figure}
We further analyze the ZC metrics designed by our framework.
As shown in Fig.~\ref{fig:freq_1}, each hand-crafted ZC metric is used at least once in the synthesized metrics.
Notably, the \textbf{MeCo} metric is selected by SR in all 31 runs, whereas the \textbf{SWAP} metric is only chosen once.
Although \textbf{SWAP} exhibits relatively high rank-correlation across the three experimental problems (see Fig.~\ref{fig:corr_nb}), its combination with other hand-crafted ZC metrics appears to be less promising.
Interestingly, \textbf{Snip} is the second most chosen metric by our framework despite its modest $\tau$ scores across the three experimental problems.
These findings suggest that our framework does not merely favor metrics with high rank-correlation but instead prioritizes those metrics that contribute most effectively when combined with others.

Among 31 final ZC metrics returned by our framework, we evaluate Kendall's $\tau$ for each problem and compute the score for each candidate according to Equation \ref{eqn:score}.
The metric with the highest score is selected for use and comparison with other ZC metrics in later experiments.
This ZC metric is presented in Equation \ref{eqn:best_expression}, which combines six hand-crafted ZC metrics: \textbf{FLOPs}, \textbf{Snip}, \textbf{L2-norm}, \textbf{Zen}, \textbf{ZiCo}, and \textbf{MeCo}.
\begin{strip}
\begin{align}\label{eqn:best_expression}
{
    \frac{ZiCo \times MeCo^2 \times log(FLOPs)}{(MeCo + Zen) \times (\sqrt{Snip} \times (MeCo + Zen + 2\times L2\text{-}norm) + MeCo)}
}
\end{align}
\end{strip}

\subsection{Insights from analyzing the synthesized ZC metric and the search process of our SR framework}
The ZC metrics used as input features for symbolic regression (SR) in this study can be categorized into two types: data-agnostic metrics (e.g., L2-Norm, Params, Synflow, and Zen) and data-dependent metrics (e.g., EPE-NAS, Fisher, FLOPs, Grad-norm, Grasp, Jacov, NWOT, Plain, SNIP, ZiCo, MeCo, and SWAP).
In this section, we assess the impact of each type on the performance of our framework by comparing the best combination of data-agnostic metrics, the best combination of data-dependent metrics, and our synthesized metric (Equation \ref{eqn:best_expression}), which integrates both types.
We note that the input dataset for SR comprises the NAS problems NB101-CF10, NB201-CF10, and NB301-CF10.
\begin{table*}[!hb]
\caption{Comparisons to the best combination of data-agnostic ZC metrics and the best combination of data-dependent ZC metrics for the NB101-CF10, NB201-CF10, NB301-CF10, and TNB101-Micro-Scene. Best and worst results are presented in \textcolor{red}{\textbf{red}} and \textcolor{blue}{\textbf{blue}}, respectively.}\label{tab:compare_agnostic_dependent}
\setstretch{1.15}
\centering
{
\begin{tabular}{lcccc}
\toprule
ZC Metric                  & NB101-CF10 & NB201-CF10 & NB301-CF10 & TNB101-Micro-Scene \\ \midrule
Data-Agnostic     & \textcolor{blue}{\textbf{0.41}} & \textcolor{blue}{\textbf{0.60}} & \textcolor{blue}{\textbf{0.31}} & 0.41 \\
Data-Dependent & 0.44 & 0.71 & 0.39 & \textcolor{blue}{\textbf{0.20}} \\
\textbf{Ours (Equation~\ref{eqn:best_expression})}    & \textcolor{red}{\textbf{0.61}} & \textcolor{red}{\textbf{0.76}} & \textcolor{red}{\textbf{0.40}} & \textcolor{red}{\textbf{0.48}} \\
\bottomrule
\end{tabular}
}
\end{table*}
{As shown in Table~\ref{tab:compare_agnostic_dependent}, the combination of data-dependent metrics is better than the combination of data-agnostic ones for these problems included in the input dataset (i.e., NB101-CF10, NB201-CF10, and NB301-CF10).
However, when applied to TNB101-Micro-Scene (which is the problem outside the input dataset), the data-agnostic metrics demonstrate better performance.
This suggests that relying solely on data-dependent metrics may lead to overfitting to the problems in the input dataset, reducing their effectiveness when applied to unseen problems.
Conversely, while data-agnostic metrics show strong generalizability across diverse problems, their overall performance is less impressive.
By simultaneously incorporating both types of metrics, our synthesized metric achieves not only superior performance on problems within the input dataset but also demonstrates high generalizability and strong performance across a variety of unseen problems.}

\begin{equation}\label{eqn:first_expression}
    f(\cdot) = \frac{ZiCo}{L2\text{-}norm} \times \sqrt{MeCo}
\end{equation}
We also obtain some interesting findings when comparing the best expression tree in the first generation (Equation~\ref{eqn:first_expression}) to the best one in the final generation (Equation~\ref{eqn:best_expression}).
First, there is a noticeable increase in complexity, with the number of metrics rising from 3 (ZiCo, L2-norm, MeCo) to 6 (ZiCo, MeCo, Zen, SNIP, L2-norm, FLOPs).
We suppose that this increased complexity enables the synthesized metric to capture more characteristics of architectures.
Second, the core metrics ZiCo, L2-norm, and MeCo are consistently retained across both versions, demonstrating that the genetic programming process effectively identifies and preserves metrics that contribute significantly to high performance.
Lastly, the increased complexity results in substantial improvements in Kendall's Tau scores, indicating a stronger correlation with true performance: from 0.39, 0.70, and 0.35 (for the initial synthesized metric) to 0.61, 0.76, and 0.40 (for the final synthesized metric) across three NAS problems.

\begin{table}[!tbh]
\caption{Kendall's Tau score of our best synthesized ZC metric (Equation~\ref{eqn:best_expression}) when replacing the most frequently used ZC metrics (i.e., ZiCo, Snip, and MeCo) with the least frequently used ZC metrics (SWAP, Grasp, and Plain).
The results are colored \textcolor{blue}{\textbf{blue}} if the replacement causes a decreasing in performance.}\label{tab:replace_zc_metrics}
\setstretch{1.15}
\centering
\resizebox{\linewidth}{!}
{
\begin{tabular}{lcccc}
\toprule
Replacing                  & NB101-CF10 & NB201-CF10 & NB301-CF10  \\ \midrule
MeCo $\rightarrow$ SWAP     & \textcolor{blue}{\textbf{0.52}} & \textcolor{blue}{\textbf{0.41}} & \textcolor{blue}{\textbf{0.39}} \\
ZiCo $\rightarrow$ Grasp     & \textcolor{blue}{\textbf{0.15}} & \textcolor{blue}{\textbf{0.45}} & \textcolor{blue}{\textbf{0.22}} \\
SNIP $\rightarrow$ Plain     & \textcolor{blue}{\textbf{0.43}} & \textcolor{blue}{\textbf{0.71}} & \textcolor{blue}{\textbf{0.33}} \\
\midrule
\textbf{Ours (Eqn.~\ref{eqn:best_expression})}    & \textbf{0.61} & \textbf{0.76} & \textbf{0.40} \\
\bottomrule
\end{tabular}
}
\end{table}
We also explore the impact of replacing the most frequently used ZC metrics in our synthesized metric (e.g., ZiCo, Snip, and MeCo in Equation~\ref{eqn:best_expression}) with the least frequently used ones (e.g., SWAP, Grasp, and Plain) and compare the Kendall's Tau scores of these two variants.
As shown in Table~\ref{tab:replace_zc_metrics}, the effectiveness of the synthesized metric is significantly reduced in all replacement cases.
This result, coupled with the presence of effective metrics like ZiCo and MeCo in both the initial and final populations, demonstrates that our SR framework effectively identifies these metrics as crucial components when combined with others to form potential ``building blocks.'' 
The genetic programming process then assembles these building blocks to create high-performing ZC metrics.
Therefore, substituting components within these building blocks disrupts their structures, leading to a noticeable decline in the performance of the synthesized metric.

\subsection{Comparison to other ZC metrics on NAS-Bench-Suite-Zero}\label{ssec:result_2}
\begin{figure*}[tbh]
    \centering
    \includegraphics[width=0.8\linewidth]{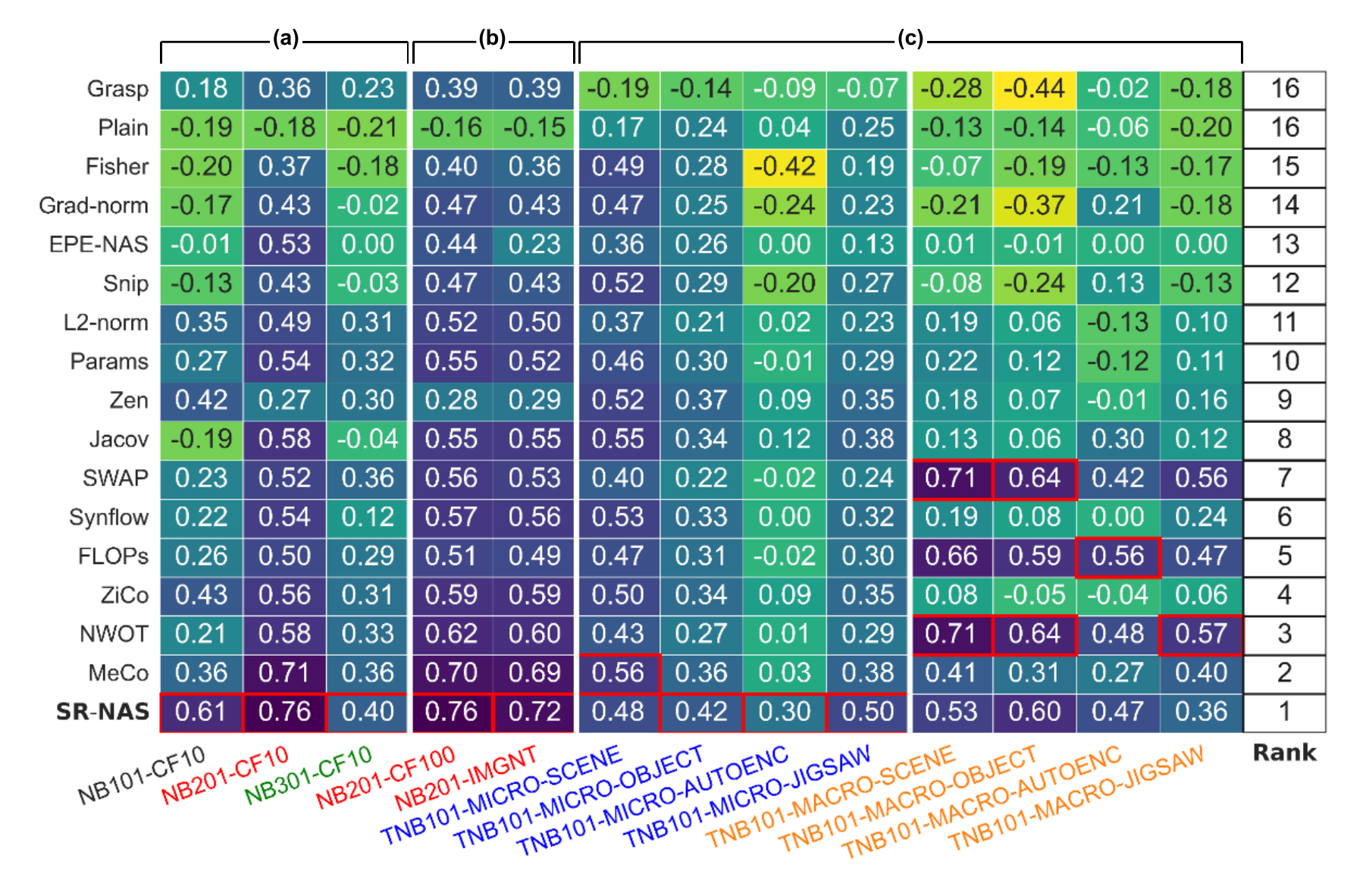}
    \caption{Kendall's $\tau$ rank correlation between ZC metrics values and networks' ground truth performance. The rows are arranged based on the average rank of metrics across all problems, with problems within the same search space colored identically. The highest Kendall's $\tau$ for each problem is highlighted with a red box. The results are presented in three groups: \textbf{(a)} Problems \emph{included} in the dataset used by our framework in the search procedure; \textbf{(b)} Problems within the same search space as one of the problems in (a), but with different tasks; \textbf{(c)} Problems \emph{not included} in the dataset used by our framework in the search procedure. Our ZC metric is denoted as \textbf{SR-NAS}.}
    \label{fig:corr_nb}
\end{figure*}
\textbf{Comparison to hand-crafted ZC metrics:} Fig.~\ref{fig:corr_nb} exhibits the Kendall's $\tau$ score comparisons between our obtained ZC metric (i.e., Equation~\ref{eqn:best_expression}) (denoted as \textbf{SR-NAS}) and hand-crafted ZC metrics across 13 problems in NAS-Bench-Suite-Zero.
We additionally visualize the correlation between our ZC metric and the ground-truth performance in Appendix H of the Supplementary. 
The results indicate that Kendall's $\tau$ scores of our ZC metric significantly surpass those of other metrics for the problems used to build the input dataset of our framework (i.e., NB101-CF10, NB201-CF10, and NB301-CF10). For the NB201-CF100 and NB201-IMGNT problems, which reside within the same search space as one of the problems in the input dataset (i.e., NB201-CF10), our ZC metric also achieves the highest Kendall's $\tau$ scores.
We further assess the generalizability of our metric by validating it on unseen search spaces (i.e., TransNAS-Bench-101 Micro/Macro) and tasks (i.e., Object detection, Scene classification, Jigsaw puzzle, and Autoencoding).
As depicted in Fig.~\ref{fig:corr_nb}, our metric yields the highest scores in 3 out of 4 problems in the TransNAS-Bench-101-Micro and achieves comparable scores for the remaining problems. 
Additionally, we rank the ZC metrics based on the sum of their rankings across all problems.
Fig.~\ref{fig:corr_nb} shows that \textbf{SR-NAS} holds the top rank, underscoring the impressive stability of our ZC metric across various problems.
These results demonstrate the effectiveness of leveraging hand-crafted ZC metrics to automatically synthesize a robust ZC metric that performs consistently well across multiple search spaces and tasks.

\textbf{Comparison to other automatic frameworks:}
\begin{table*}[!tbh]
\caption{Comparisons in Kendall's $\tau$ rank correlation between frameworks that automate the design of ZC metrics. The symbol `-' indicates that the results were not reported in the original studies.
Best results are presented in \textcolor{red}{\textbf{red}}.
We do not compare to Auto-Prox~\cite{autoprox} because it serves for vision transformer networks and thus could not test on NAS-Bench-Suite-Zero.
}\label{tab:compare_2_auto}
\setstretch{1.15}
\centering
{
\begin{tabular}{lccccc}
\toprule
Framework & NB201-CF10 & NB201-CF100 & NB201-IMGNT & Search Cost (hours)\\ \midrule
EZNAS \cite{ez_nas} & 0.65 & 0.65 & 0.61 & 24.0 \\
UP-NAS \cite{up_nas} & 0.71 & - & - & 0.03 \\
RoBoT \cite{robot} & 0.55 & 0.57 & 0.57 & - \\
AZ-NAS \cite{az_nas} & 0.73 & 0.72 & 0.69 & - \\ 
\midrule
SR-NAS (Ours) & \textcolor{red}{\textbf{0.76}} & \textcolor{red}{\textbf{0.76}} & \textcolor{red}{\textbf{0.72}} & 0.17\\ \bottomrule
\end{tabular}
}
\end{table*}
Table~\ref{tab:compare_2_auto} presents a comparison of both Kendall's $\tau$ scores and search costs between our framework and other frameworks that also automate the design of ZC metrics (i.e., EZNAS, UP-NAS, RoBot, and AZ-NAS).
The results indicate that our metric consistently yields the highest Kendall's $\tau$ scores across all three problems in NAS-Bench-201.
Notably, our framework outperforms EZNAS, even though both frameworks utilize Symbolic Regression (SR) as the search algorithm.
This suggests the effectiveness of using high-level features (i.e., hand-crafted ZC metrics) over low-level features (i.e., network statistics) to design a robust ZC metric.
On the other hand, the benefit of using SR in synthesizing ZC metrics is highlighted by our higher $\tau$ scores compared to UP-NAS and RoBoT, despite these frameworks also using hand-crafted ZC metrics as input features.

In terms of search cost, our framework (0.17 hours $\approx$ 10 minutes) and UP-NAS (0.03 hours $\approx$ 2 minutes) are significantly lower than EZ-NAS (24 hours) due to leveraging the NAS benchmarks.
Although our time is slightly higher than UP-NAS (i.e., 8 minutes), it is important to note that our framework only requires a single search, and our obtained ZC metric can be employed as an effective proxy across different NAS problems.
In contrast, UP-NAS, RoBot and AZ-NAS require re-conducting the search process for each specific problem.

\subsection{Capability of searching top-performing networks}
\begin{table}[tbh]
\caption{{Kendall's Tau rank correlation and test accuracy on ImageNet of the best networks found by ZC metrics within OFA search spaces.
Best results are presented in \textcolor{red}{\textbf{red}}.}}\label{tab:compare_ofa}
\setstretch{1.15}
\centering
{
\begin{tabular}{lcc}
\toprule
Metric                  & Kendall's Tau  & \begin{tabular}[c]{@{}c@{}}Top-1 Accuracy\\(\%)\end{tabular} \\ \midrule
MeCo \cite{meco}       & 0.51 & 76.40 \\
Zen \cite{zen}         & 0.59 & 76.43 \\
FLOPs                   & 0.60 & 76.30 \\
Snip \cite{snip}       & 0.04 & 76.13 \\
L2-norm \cite{zc_nas}  & 0.41 & 76.51 \\
ZiCo \cite{zico}       & \textcolor{red}{\textbf{0.68}} & 76.44 \\
\midrule
\textbf{Ours}           & 0.60 & \textcolor{red}{\textbf{76.65}} \\ \midrule
\textit{\begin{tabular}[c]{@{}l@{}}Optimal\\(in 1000 networks)\end{tabular}}        & - & \textit{76.81} \\
\bottomrule
\end{tabular}
}
\end{table}

\begin{table*}[tbh]
\caption{Comparison of error rates (\%) on the CIFAR-10 dataset for networks discovered in the DARTS search space.
For training-free methods, the specific ZC metrics used by the algorithms are listed in parentheses. The search cost is measured in GPU days.
}\label{tab:darts_res}
\setstretch{1.15}
\begin{center}
{
\begin{tabular}{lccccc}
\toprule
Method            & Test Error (\%) & Params (M) & Search Cost & Training-free  \\ \midrule
PNAS~\cite{pnas} & $3.41\pm0.09$ & 3.2 & 225 \\
DARTS~\cite{darts}  & $3.00\pm0.14$ & 3.3 & 4.0  \\
RandomNAS~\cite{randomnas}  & $2.85\pm0.08$ & 4.3 & 2.7 \\
DARTS-PT~\cite{darts_pt}  & $2.61\pm0.08$ & 3.0 & 0.8  \\
PreNAS~\cite{prenas} & $2.49\pm0.09$ & 4.5 & 0.6 \\
PINAT~\cite{pinat} & $2.54\pm0.08$ & 3.6 & 0.3 \\ \midrule
TE-NAS (NLR, NTK)~\cite{tenas} & $2.63\pm0.06$ & 3.8 & 0.03 & \cmark \\
Zero-Cost-PT (ZiCo)~\cite{zico} & $2.80\pm0.03$ & 5.1 & 0.04 & \cmark \\
Zero-Cost-PT (MeCo)~\cite{meco} & $2.69\pm0.05$ & 4.2 & 0.08 & \cmark \\
SWAP-NAS$^\dag$ (SWAP)~\cite{swap} & $2.48\pm0.09$ & 4.3 & 0.004 & \cmark \\ \midrule
Aging Evolution (SR-NAS) (Ours)  & $2.66\pm0.04$ & 3.9 & 0.01 & \cmark \\
\bottomrule
\end{tabular}
}
\end{center}
\footnotesize{$^\dag$ SWAP-NAS searches in a sub-space of DARTS, where the normal and reduction cells are similar.}
\end{table*}
In addition to the rank correlation with network performance, the ability to identify top-performing networks is a crucial aspect of ZC metrics.
Previous studies have highlighted the limitations of ZC metrics in effectively identifying top-performing networks, even when they achieve high rank-correlation scores~\cite{meco, gecco_24}.
{We verify this ability of our obtained ZC metric by testing it on the large-scale ImageNet dataset. Specifically, we randomly sample 1,000 networks from the Once-For-All (OFA~\cite{ofa}) search space and then compute the rank correlation between our ZC metric score and the networks' accuracy on the ImageNet dataset. Table~\ref{tab:compare_ofa} reveal that our Kendall rank correlation is only slightly lower than ZiCo (i.e., 0.60 compared to 0.68). However, the network with the highest score according to our ZC metric achieves the highest top-1 accuracy among all networks identified by the competing ZC metrics. This demonstrates the ability of our synthesized metric to reliably identify high-performing networks in large-scale search spaces such as OFA.}
We further employ our metric as the search objective of Aging Evolution~\cite{rea}, which is a widely-used evolutionary algorithm for NAS~\cite{rea, swap, freerea}.
The algorithm is deployed in the DARTS search space~\cite{darts} to seek high-performance networks for the CIFAR-10 dataset.
As shown in Table \ref{tab:darts_res}, the algorithm using our SR-designed ZC metric as the search objective could figure out a network with comparable performance within 15 minutes (i.e., 0.01 GPU days) (the network architecture found by our algorithm is exhibited in Appendix F of the Supplementary).
This result demonstrates both the efficiency and the effectiveness of our obtained ZC metric in discovering high-performance networks.
We additionally provide a comparison of the best networks identified by our ZC metric versus other metrics across all problems in NAS-Bench-Suite-Zero in Appendix G of the Supplementary, showing that the networks found with our metric are the best compared to those found with hand-crafted ZC metrics in 8 out of 13 total problems.

\subsection{Ablation Study}\label{ssec:ablation_study}
\textbf{Extensibility:}
We demonstrate the extensibility of our proposed framework by additionally experimenting with \emph{two} variants of the current input dataset (details of each dataset are presented in Appendix B of the Supplementary).
The results detailed in Appendix B demonstrate that augmenting the input dataset with robust ZC metrics can further enhance the performance of our proposed framework.
Additionally, we hypothesize that the generalizability of our framework could be improved by using a dataset that is augmented with a broader range of problems.
Moreover, the metric found by the framework using the dataset covering more problems delivers better general performance across 13 problems in NAS-Bench-Suite-Zero than the current best metric (i.e., Equation \ref{eqn:best_expression}).

\textbf{Lack of generalizability when searching on a single problem:}
We conduct additional experiments to investigate the potential overfitting of our framework when searching with the input dataset that only covers a single problem.
Table~\ref{tab:abl_overfit} reveals that the $\tau$ scores of ZC metrics found by using the datasets that cover only a single problem significantly deteriorate when evaluated on other problems.
For instance, the metric found by using the dataset built on the NB301-CF10 problem achieves a score of 0.48 for NB301-CF10 but has a negative score of -0.17 for NB201-CF10.
Conversely, the ZC metric found by using the dataset containing three problems achieves a more balanced performance across all problems.
\begin{table}[!ht]
\caption{Comparisons in Kendall's $\tau$ rank correlations of the best ZC metrics (over 31 runs) designed by our framework using the input datasets that only cover a single problem (i.e., NB101-CF10, NB201-CF10, and NB301-CF10) and using the input dataset containing all 3 problems.
Best results are presented in \textcolor{red}{\textbf{red}} color.}\label{tab:abl_overfit}
\setstretch{1.15}
\centering
\resizebox{\linewidth}{!}
{
\begin{tabular}{lcccc}
\toprule
Search Problem & NB101-CF10 & NB201-CF10 & NB301-CF10 \\ \midrule
NB101-CF10  & \textcolor{red}{\textbf{0.66}} & 0.70 & 0.37 \\
NB201-CF10  & 0.47 & \textcolor{red}{\textbf{0.79}} & 0.27 \\
NB301-CF10  & 0.36 & -0.17 & \textcolor{red}{\textbf{0.48}} \\ \midrule
All 3 problems         & 0.61 & 0.76 & 0.40 \\
\bottomrule
\end{tabular}
}
\end{table}

\section{Conclusion}
In this paper, we propose a novel framework that performs Symbolic Regression (SR) via genetic programming to automatically design robust zero-cost (ZC) metrics for various NAS search spaces and tasks.
Our approach introduces several notable features.
First, unlike previous frameworks, the input dataset of our framework comprises diverse problems instead of a single one.
We further introduce a novel evaluation mechanism that assesses the quality of generated ZC metrics throughout the search process.
This mechanism not only guides the genetic programming process to identify expressions that correlate well with network performance across multiple problems but also prevents it from overfitting to a specific problem.
Additionally, our framework is highly extensible; incorporating more powerful hand-crafted ZC metrics and diverse problems into the input dataset could enhance the effectiveness of resulting metrics across various problems.
Extensive experiments on NAS-Bench-Suite-Zero demonstrate both the efficacy and efficiency of our method.
Our framework could design a ZC metric with state-of-the-art Kendall's $\tau$ correlation on NAS-Bench-101/201/301 search spaces within 10 minutes, and our metric remains competitive to hand-crafted metrics on TransNAS-Bench-101-Micro/Macro search spaces.
When optimizing our obtained SR-NAS metric with an evolutionary algorithm, we could figure out a network architecture with comparable performance in the DARTS search space within 15 minutes.


\bibliographystyle{IEEEtran}
\bibliography{bare_jrnl_new_sample4}


\section*{Supplementary material}

# From Hand-Crafted Metrics to Evolved Training-Free Performance Predictors for Neural Architecture Search via Genetic Programming (Supplementary Material)

## A Full results in Section IV-A

Table 1: Kendall's $\tau$ correlation (mean $\pm$ standard deviation) of the returned expressions over 31 runs on the input/test/(input + test) datasets for NB101/201/301-CF10 problems.

| Problem | Input | Test | Input + Test |
|---|---|---|---|
| NB101-CF10 | $0.5587 \pm 0.0258$ | $0.5603 \pm 0.0265$ | $0.5592 \pm 0.0255$ |
| NB201-CF10 | $0.7393 \pm 0.0221$ | $0.7369 \pm 0.0228$ | $0.7386 \pm 0.0222$ |
| NB301-CF10 | $0.3883 \pm 0.0133$ | $0.3864 \pm 0.0134$ | $0.3877 \pm 0.0121$ |

## B Extensibility of the proposed framework

Table 2: Kendall's $\tau$ rank correlations of the best ZC metrics (over 31 runs) designed by our framework with using three different input datasets: $\mathcal{D}$, $\mathcal{D}^-$, and $\mathcal{D}^+$. Best results for each problem are presented in red color.

| Dataset | NB101-CF10 | NB201-CF10 | NB301-CF10 | TNB101-Micro-Scene | TNB101-Macro-Scene |
|---|---|---|---|---|---|
| $\mathcal{D}^-$ | 0.54 | 0.72 | 0.39 | 0.19 | -0.30 |
| $\mathcal{D}^+$ | 0.48 | **0.78** | **0.42** | **0.63** | **0.67** |
| $\mathcal{D}$ | **0.61** | 0.76 | 0.40 | 0.48 | 0.53 |

In this section, we present the full results of experiments to verify the extensibility of our proposed framework (i.e., Section IV-E). Specifically, we experiment with three following input datasets:

1. $\mathcal{D}$: The dataset used in Section IV-A to search a robust ZC metric across multiple problems. This dataset contains the scores of 16 hand-crafted ZC metrics (including 13 ZC metrics in NAS-Bench-Suite-Zero and three newly-proposed metrics (i.e., ZiCo, MeCo, SWAP)) on **NB101-CF10**, **NB201-CF10**, and **NB301-CF10** problems.
2. $\mathcal{D}^-$: We remove the scores of three novel hand-crafted ZC metrics (i.e., ZiCo, MeCo, and SWAP) from $\mathcal{D}$.
3. $\mathcal{D}^+$: We integrate into $\mathcal{D}$ the ZC metrics scores and networks performance on two additional problems: **TNB101-Micro-Scene** and **TNB101-Macro-Scene**.

For each dataset, we execute 31 independent runs with the settings as presented in Section IV-A. Table 2 exhibits the Kendall's $\tau$ scores of the best ZC metrics designed by our framework with $\mathcal{D}$, $\mathcal{D}^-$, and $\mathcal{D}^+$ input datasets. The results demonstrate that using dataset $\mathcal{D}$ achieves higher Kendall's $\tau$ scores across all testing problems than using dataset $\mathcal{D}^-$. This suggests that augmenting the input dataset with robust ZC metrics can further enhance the performance of our proposed framework. Additionally, we hypothesize that the generalizability of our framework could be improved by using a dataset that is augmented with a broader range of problems. While the $\tau$ score for the NB101-CF10 problem slightly decreases when using dataset $\mathcal{D}^+$, the $\tau$ scores or the remaining problems show an increase compared to the framework using dataset $\mathcal{D}$. Moreover, the metric found by the framework using $\mathcal{D}^+$ delivers better general performance across all 13 problems in NAS-Bench-Suite-Zero than the current best metric (i.e., Equation 2), which is found by using $\mathcal{D}$.

## C MATHEMATICAL OPERATORS

In our experiments, we implement the primitive mathematical operators for SR: {*add*, *sub*, *mul*, *div*, *neg*, *log*, *sqrt*}. The details of each operator are presented in Table 3. Among these operators, three operators have constraints in these input values: *div* (i.e., $Y \neq 0$), *log* (i.e., $X > 0$), and *sqrt* (i.e., $X \geq 0$). We return the value of -10,000,000 for these operators in cases where the input values violate constraints.

Table 3: List of mathematical operators

| Name | Operation | Type | Description |
|---|---|---|---|
| *add* | Addition | Binary | $Z = X + Y$ |
| *sub* | Subtraction | Binary | $Z = X - Y$ |
| *mul* | Multiplication | Binary | $Z = X \times Y$ |
| *div* | Division | Binary | $Z = X \div Y$ |
| *log* | Logarithm | Unary | $Z = \log(X)$ |
| *sqrt* | Square root | Unary | $Z = \sqrt{X}$ |
| *neg* | Negative | Unary | $Z = -X$ |

## D LIST OF PROBLEMS

Table 4: NAS problem description

| Problem | Search Space | Task |
|---|---|---|
| NB101-CF10 | NAS-Bench-101 | Image classification on CIFAR-10 dataset |
| NB201-CF10 | NAS-Bench-201 | Image classification on CIFAR-10 dataset |
| NB201-CF100 | | Image classification on CIFAR-100 dataset |
| NB201-IMGNT | | Image classification on ImageNet16-120 dataset |
| NB301-CF10 | NAS-Bench-301 | Image classification on CIFAR-10 dataset |
| TNB101-Micro-Scene | TransNAS-Bench-101-Micro | Scene classification on Taskonomy dataset |
| TNB101-Micro-Object | | Object detection on Taskonomy dataset |
| TNB101-Micro-AutoEnc | | Autoencoding on Taskonomy dataset |
| TNB101-Micro-Jigsaw | | Jigsaw puzzle on Taskonomy dataset |
| TNB101-Macro-Scene | TransNAS-Bench-101-Macro | Scene classification on Taskonomy dataset |
| TNB101-Macro-Object | | Object detection on Taskonomy dataset |
| TNB101-Macro-AutoEnc | | Autoencoding on Taskonomy dataset |
| TNB101-Macro-Jigsaw | | Jigsaw puzzle on Taskonomy dataset |

# E Hyperparameter Setting

Table 5: Hyperparameters of Symbolic Regression

| Hyperparameter | Value |
| --- | --- |
| Population Size | 100 |
| Tournament Size | 2 |
| Probability of Crossover | 0.7 |
| Probability of Subtree Mutation | 0.1 |
| Probability of Hoist Mutation | 0.5 |
| Probability of Point Mutation | 0.1 |
| Maximum number of generations | 50 |
| Minimum depth of expression tree | 2 |
| Maximum depth of expression tree | 10 |

Table 6: Hyperparameters of Aging Evolution

| Hyperparameter | Value |
| --- | --- |
| Population Size | 100 |
| Probability of Mutation | 0.5 |
| Maximum number of evaluations | 3000 |

# F NETWORK ARCHITECTURE FOUND IN DARTS SEARCH SPACE

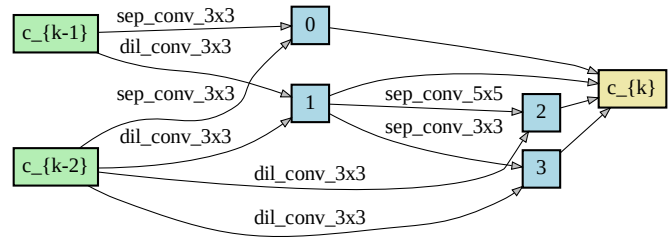

(a) Normal cell

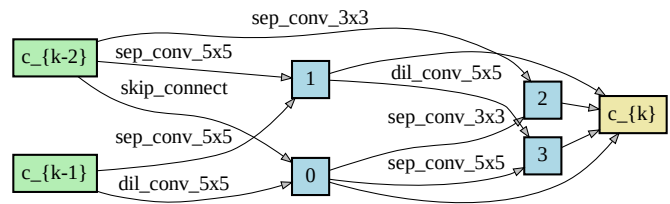

(b) Reduction cell

Figure 1: Normal and reduction cells of the architecture found by our SR-designed ZC metric in the DARTS search space for the CIFAR-10 dataset. The model size of this network is 3.97 MB.

# G PERFORMANCE COMPARISONS OF THE BEST NETWORKS FOUND BY ZC METRICS FOR PROBLEMS IN NAS-BENCH-SUITE-ZERO

Table 7: Performance comparisons of the best networks found by ZC metrics for problems within NAS-Bench-101, NAS-Bench-201, and NAS-Bench-301 search spaces. Best results for each problem are presented in red color.

| Metric | NB101-CF10 ↑ | NB201-CF10 ↑ | NB201-CF100 ↑ | NB201-IMGNT ↑ | NB301-CF10 ↑ |
|---|---|---|---|---|---|
| Params | 89.11 | 90.36 | 71.34 | 41.23 | 91.98 |
| FLOPs | 89.11 | 90.36 | 71.34 | 41.23 | 91.98 |
| Snip (Lee et al., 2019) | 86.78 | 81.45 | 47.62 | 23.40 | 91.98 |
| Fisher (Turner et al., 2020) | 86.78 | 81.45 | 47.62 | 25.93 | 91.09 |
| Jacov (Mellor et al., 2020) | 84.98 | 89.03 | 69.78 | 43.50 | 92.12 |
| Synflow (Tanaka et al., 2020) | 89.11 | 90.36 | 71.34 | 41.23 | 93.79 |
| Grasp (Wang et al., 2020)] | 86.21 | 81.45 | 47.62 | 27.60 | 92.91 |
| Plain (Abdelfattah et al., 2021) | 86.69 | 85.21 | 55.54 | 17.70 | 91.81 |
| Grad-norm (Abdelfattah et al., 2021) | 86.78 | 81.45 | 47.62 | 20.67 | 93.34 |
| EPE-NAS (Lopes et al., 2021) | 92.00 | 85.74 | 56.36 | 42.37 | 92.69 |
| L2-norm (Abdelfattah et al., 2021) | 92.56 | 88.73 | 72.04 | 45.43 | 91.98 |
| Zen (Lin et al., 2021) | 94.22 | 86.96 | 68.26 | 40.60 | 91.98 |
| NWOT (Mellor et al., 2021) | 93.33 | 89.78 | 70.14 | 45.90 | 93.54 |
| ZiCo (Li et al., 2023) | 94.22 | 90.36 | 71.34 | 41.23 | 91.98 |
| MeCo (Jiang et al., 2023) | 93.34 | 89.86 | 70.78 | 41.23 | 91.98 |
| SWAP (Peng et al., 2024) | 92.56 | 88.28 | 69.46 | 37.63 | 93.54 |
| **Ours** | **94.58** | **91.18** | **72.72** | **46.60** | **94.57** |
| *Optimal (in the benchmark)* | *94.72* | *91.57* | *73.26* | *47.33* | *94.69* |

Table 8: Performance comparisons of the best networks found by ZC metrics for problems within TransNAS-Bench-101-Micro search space. Best results for each problem are presented in red color.

| Metric | Scene ↑ | Object ↑ | Jigsaw ↑ | AutoEnc ↑ |
|---|---|---|---|---|
| Params | 53.67 | 42.14 | 85.91 | 0.46 |
| FLOPs | 53.67 | 42.14 | 85.91 | 0.46 |
| Snip (Lee et al., 2019) | 48.72 | 36.92 | 80.32 | 0.33 |
| Fisher (Turner et al., 2020) | 48.72 | 36.92 | 83.49 | 0.00 |
| Jacov (Mellor et al., 2020) | 53.72 | 41.78 | **93.99** | 0.42 |
| Synflow (Tanaka et al., 2020) | 53.67 | 39.53 | 90.91 | 0.46 |
| Grasp (Wang et al., 2020)] | 50.19 | 31.17 | 91.08 | 0.33 |
| Plain (Abdelfattah et al., 2021) | 31.51 | 31.79 | 0.03 | 0.06 |
| Grad-norm (Abdelfattah et al., 2021) | 48.72 | 36.92 | 6.82 | 0.36 |
| EPE-NAS (Lopes et al., 2021) | 52.12 | 40.86 | 92.24 | 0.46 |
| L2-norm (Abdelfattah et al., 2021) | 53.30 | 42.74 | 91.84 | 0.45 |
| Zen (Lin et al., 2021) | 53.67 | 42.14 | 85.91 | 0.46 |
| NWOT (Mellor et al., 2021) | 53.16 | **43.02** | 92.29 | 0.41 |
| ZiCo (Li et al., 2023) | 53.67 | 42.14 | 85.91 | 0.46 |
| MeCo (Jiang et al., 2023) | 52.89 | 42.16 | 91.42 | 0.46 |
| SWAP (Peng et al., 2024) | 21.95 | 40.24 | 6.82 | 0.23 |
| **Ours** | **54.24** | 42.51 | 90.62 | **0.52** |
| *Optimal (in the benchmark)* | *54.94* | *45.59* | *95.37* | *0.58* |

Table 9: Performance comparisons of the best networks found by ZC metrics for problems within TransNAS-Bench-101-Macro search space. Best results for each problem are presented in red color.

| Metric | Scene ↑ | Object ↑ | Jigsaw ↑ | AutoEnc ↑ |
|---|---|---|---|---|
| Params | 54.26 | 42.71 | 94.26 | 0.37 |
| FLOPs | 55.52 | 45.29 | **96.51** | 0.66 |
| Snip (Lee et al., 2019) | 54.50 | 42.55 | 95.20 | 0.64 |
| Fisher (Turner et al., 2020) | 51.41 | 42.55 | 93.15 | 0.44 |
| Jacov (Mellor et al., 2020) | 56.01 | 46.05 | 95.20 | 0.65 |
| Synflow (Tanaka et al., 2020) | **56.27** | 45.29 | **96.51** | 0.41 |
| Grasp (Wang et al., 2020)] | 50.36 | 41.79 | 91.55 | 0.39 |
| Plain (Abdelfattah et al., 2021) | 51.70 | 43.33 | 92.09 | 0.41 |
| Grad-norm (Abdelfattah et al., 2021) | 53.01 | 42.55 | 93.48 | 0.64 |
| EPE-NAS (Lopes et al., 2021) | 52.61 | 45.22 | 92.93 | 0.41 |
| L2-norm (Abdelfattah et al., 2021) | 56.00 | 44.50 | 94.26 | 0.58 |
| Zen (Lin et al., 2021) | 56.00 | 44.50 | 95.35 | 0.62 |
| NWOT (Mellor et al., 2021) | 55.52 | 45.29 | **96.51** | 0.66 |
| ZiCo (Li et al., 2023) | 52.43 | 42.79 | 95.35 | 0.58 |
| MeCo (Jiang et al., 2023) | **56.27** | **46.66** | **96.51** | 0.64 |
| SWAP (Peng et al., 2024) | 55.52 | 45.29 | **96.51** | 0.66 |
| **Ours** | 54.78 | 45.74 | 96.50 | **0.73** |
| *Optimal (in the benchmark)* | *57.41* | *47.42* | *97.02* | *0.75* |

# H VISUALIZATION OF RANK CORRELATION BETWEEN OUR SR-DESIGNED ZC METRIC AND THE NETWORK PERFORMANCE

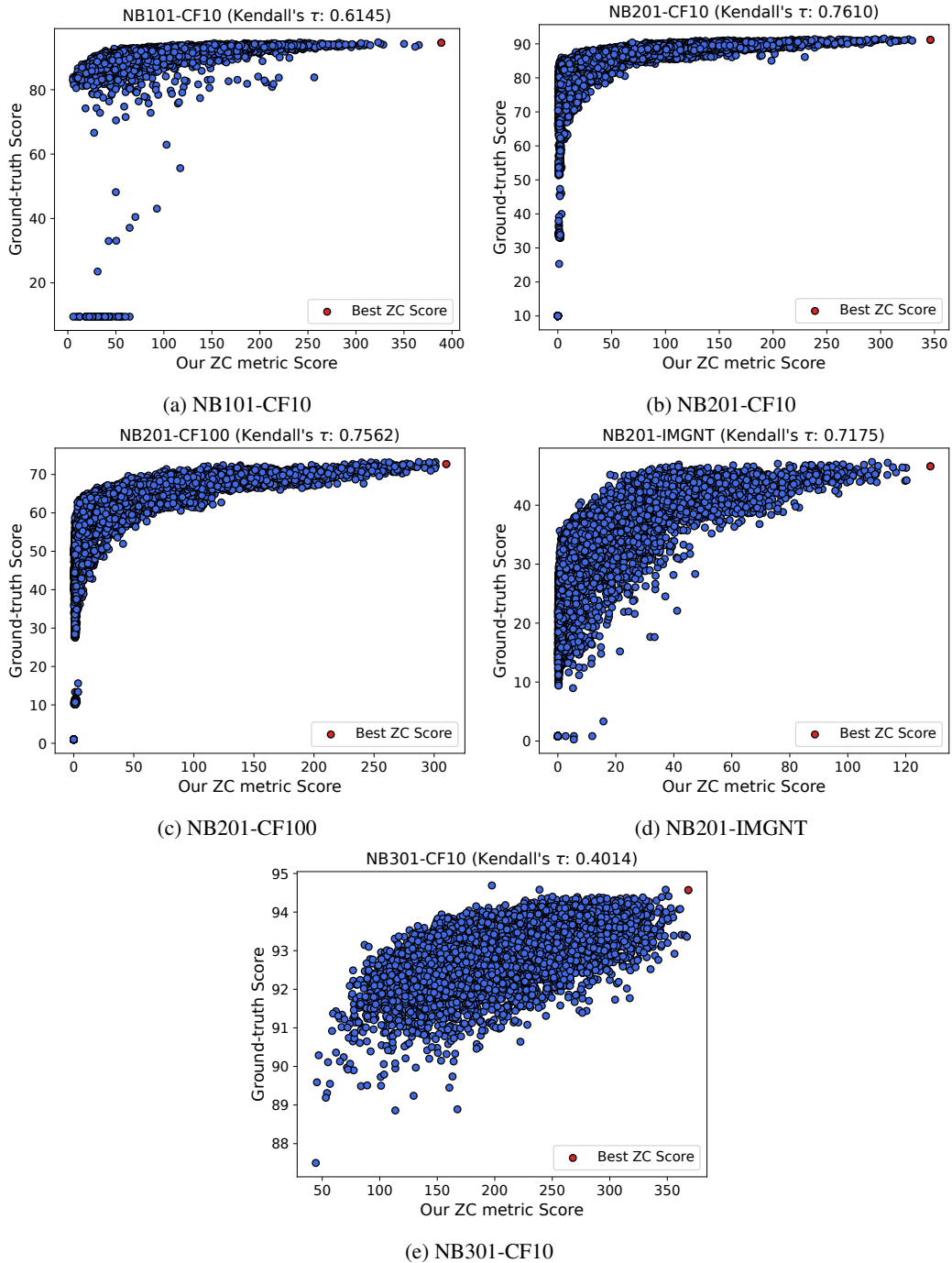

(a) NB101-CF10

(b) NB201-CF10

(c) NB201-CF100

(d) NB201-IMGNT

(e) NB301-CF10

Figure 2: Visualizations of the network performance and our ZC metric scores for the problems within the NAS-Bench-101, NAS-Bench-201, and NAS-Bench-301 search spaces.

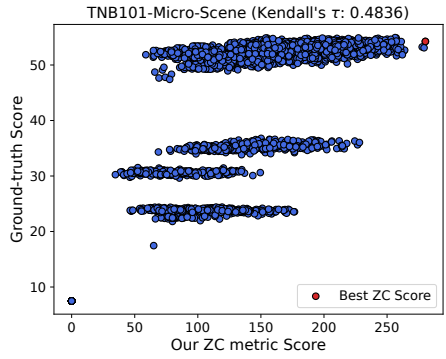
(a) TNB101-Micro-Scene

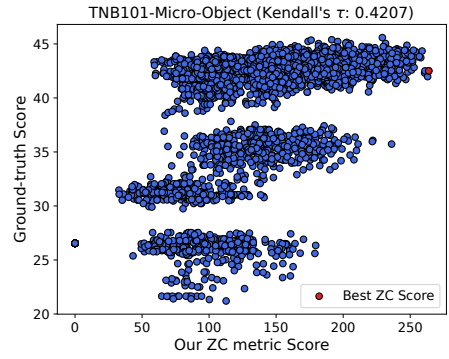
(b) TNB101-Micro-Object

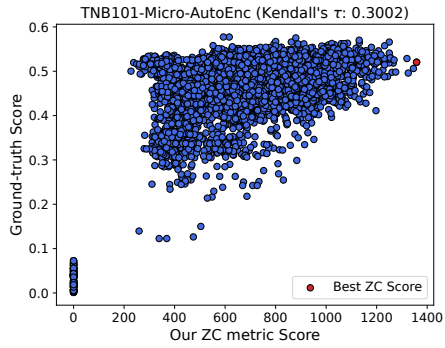
(c) TNB101-Micro-AutoEnc

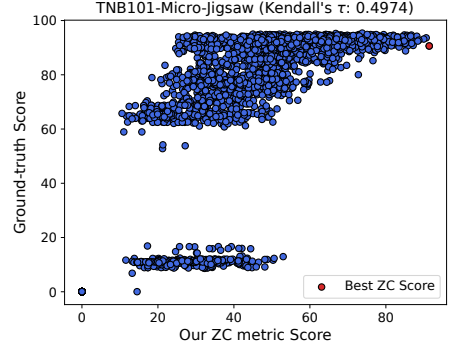
(d) TNB101-Micro-Jigsaw

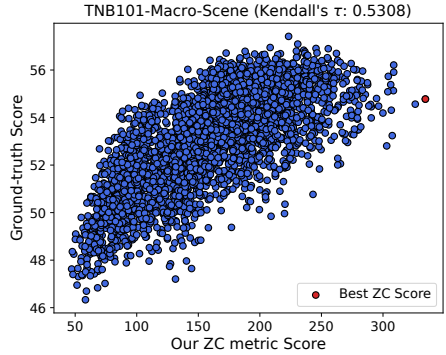
(e) TNB101-Macro-Scene

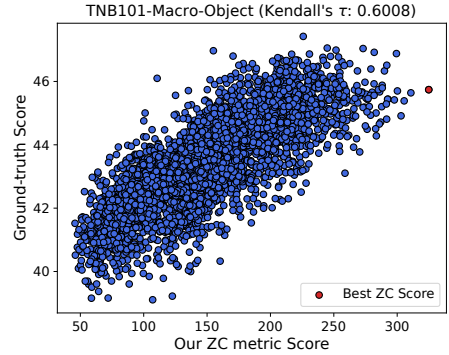
(f) TNB101-Macro-Object

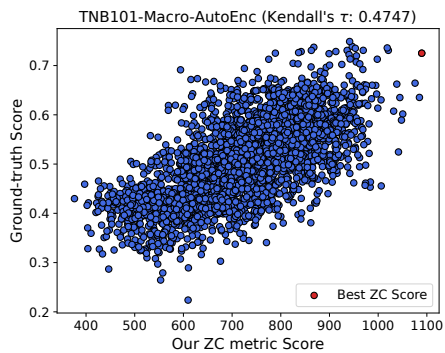
(g) TNB101-Macro-AutoEnc

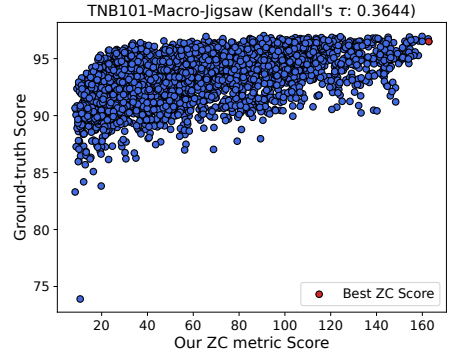
(h) TNB101-Macro-Jigsaw

Figure 3: Visualizations of the network performance and our ZC metric scores for the problems within the TransNAS-Bench-101-Micro and TransNAS-Bench-101-Macro search spaces.

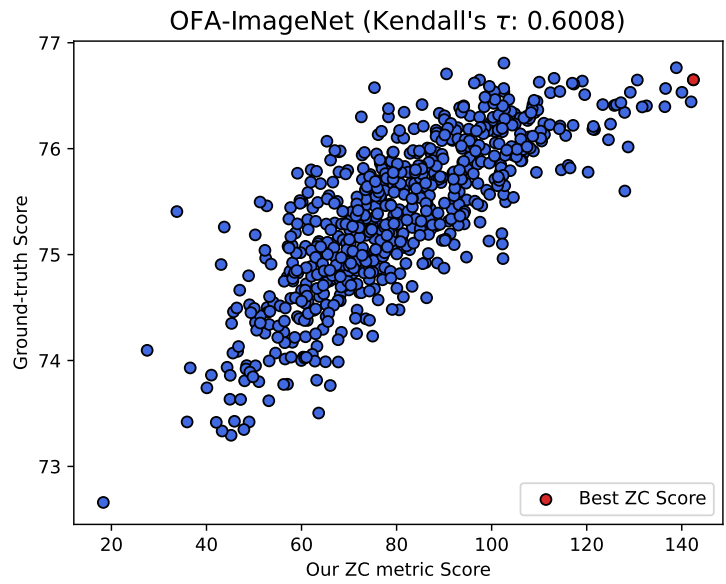

Figure 4: Visualizations of the network performance and our ZC metric scores on the ImageNet dataset for 1000 networks within the Once-For-All search space.

# I Zero-cost metrics synthesized by our framework at 31 runs

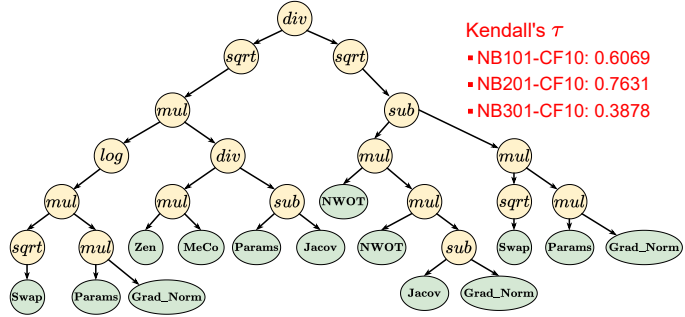

(Run 1)

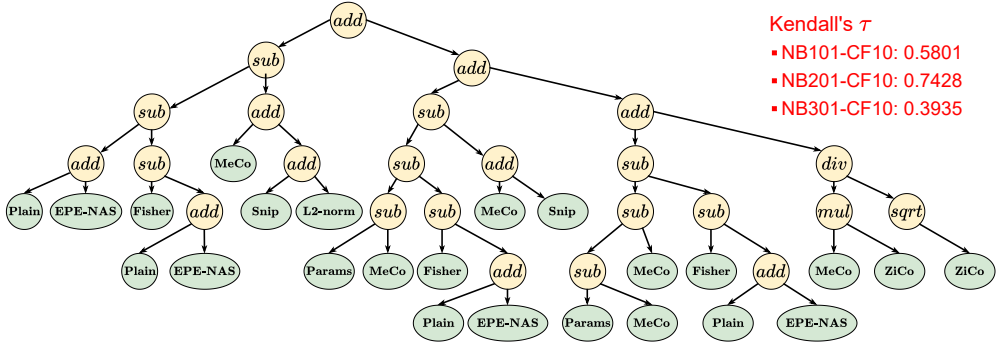

(Run 2)

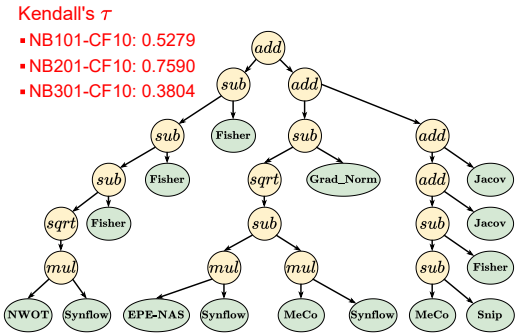

(Run 3)

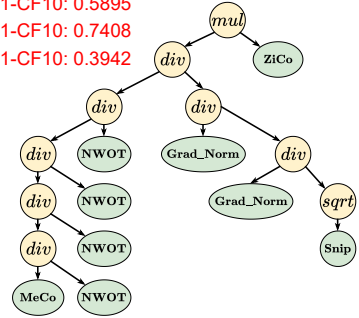

(Run 4)

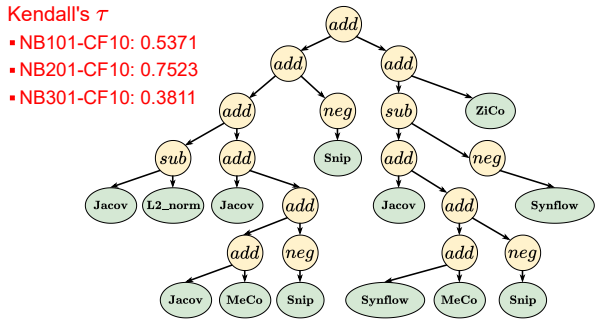

(Run 5)

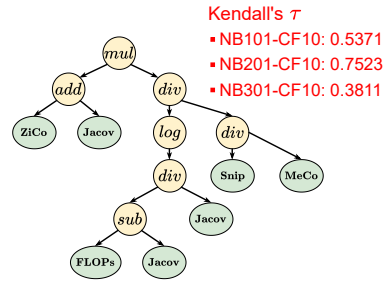

(Run 6)

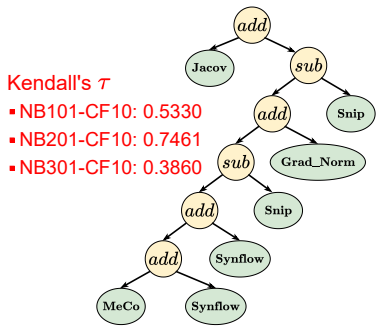

(Run 7)

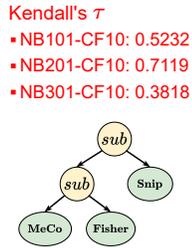

(Run 8)

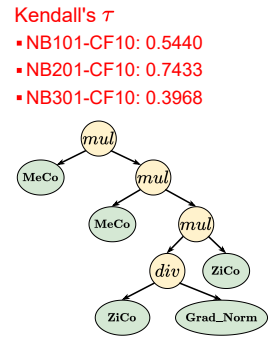

(Run 9)

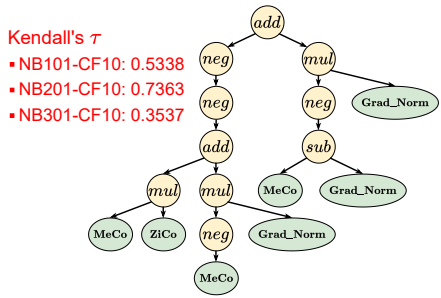

(Run 10)

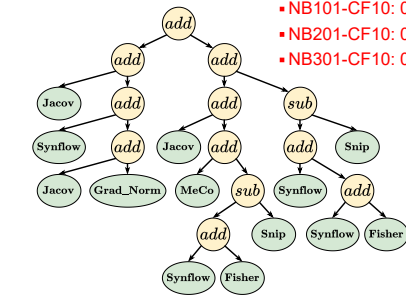

(Run 11)

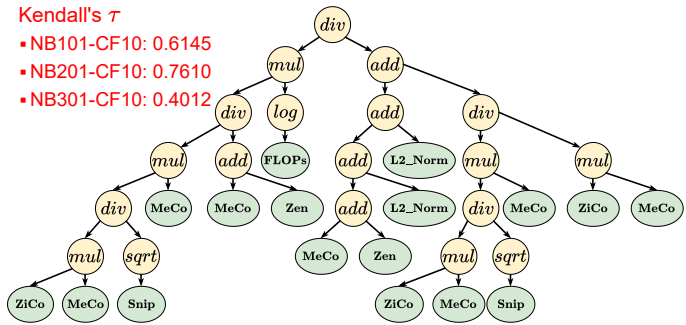

(Run 12)

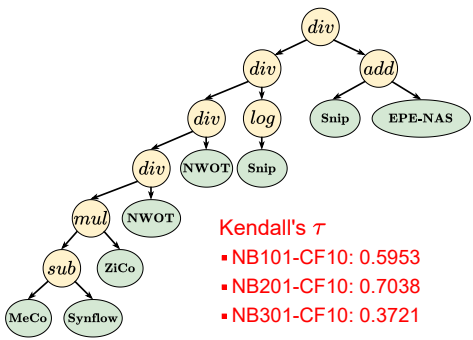

(Run 13)

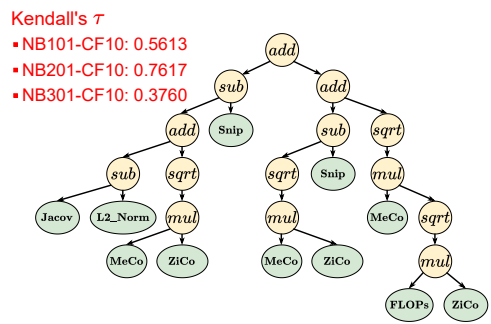

(Run 14)

(Run 15)

(Run 16)

(Run 17)

(Run 18)

(Run 19)

(Run 20)

Kendall's $\tau$
- NB101-CF10: 0.5644
- NB201-CF10: 0.7454
- NB301-CF10: 0.3929

(Run 21)

Kendall's $\tau$
- NB101-CF10: 0.5335
- NB201-CF10: 0.7358
- NB301-CF10: 0.3889

(Run 22)

Kendall's $\tau$
- NB101-CF10: 0.5778
- NB201-CF10: 0.7414
- NB301-CF10: 0.3946

(Run 23)

Kendall's $\tau$
- NB101-CF10: 0.5904
- NB201-CF10: 0.7346
- NB301-CF10: 0.4002

(Run 24)

Kendall's $\tau$
- NB101-CF10: 0.5555
- NB201-CF10: 0.6962
- NB301-CF10: 0.3984

(Run 25)

Kendall's $\tau$
- NB101-CF10: 0.5317
- NB201-CF10: 0.7397
- NB301-CF10: 0.3746

(Run 26)

Kendall's $\tau$
- NB101-CF10: 0.5595
- NB201-CF10: 0.7406
- NB301-CF10: 0.4023

(Run 27)

Kendall's $\tau$
- NB101-CF10: 0.5546
- NB201-CF10: 0.6575
- NB301-CF10: 0.4010

(Run 28)

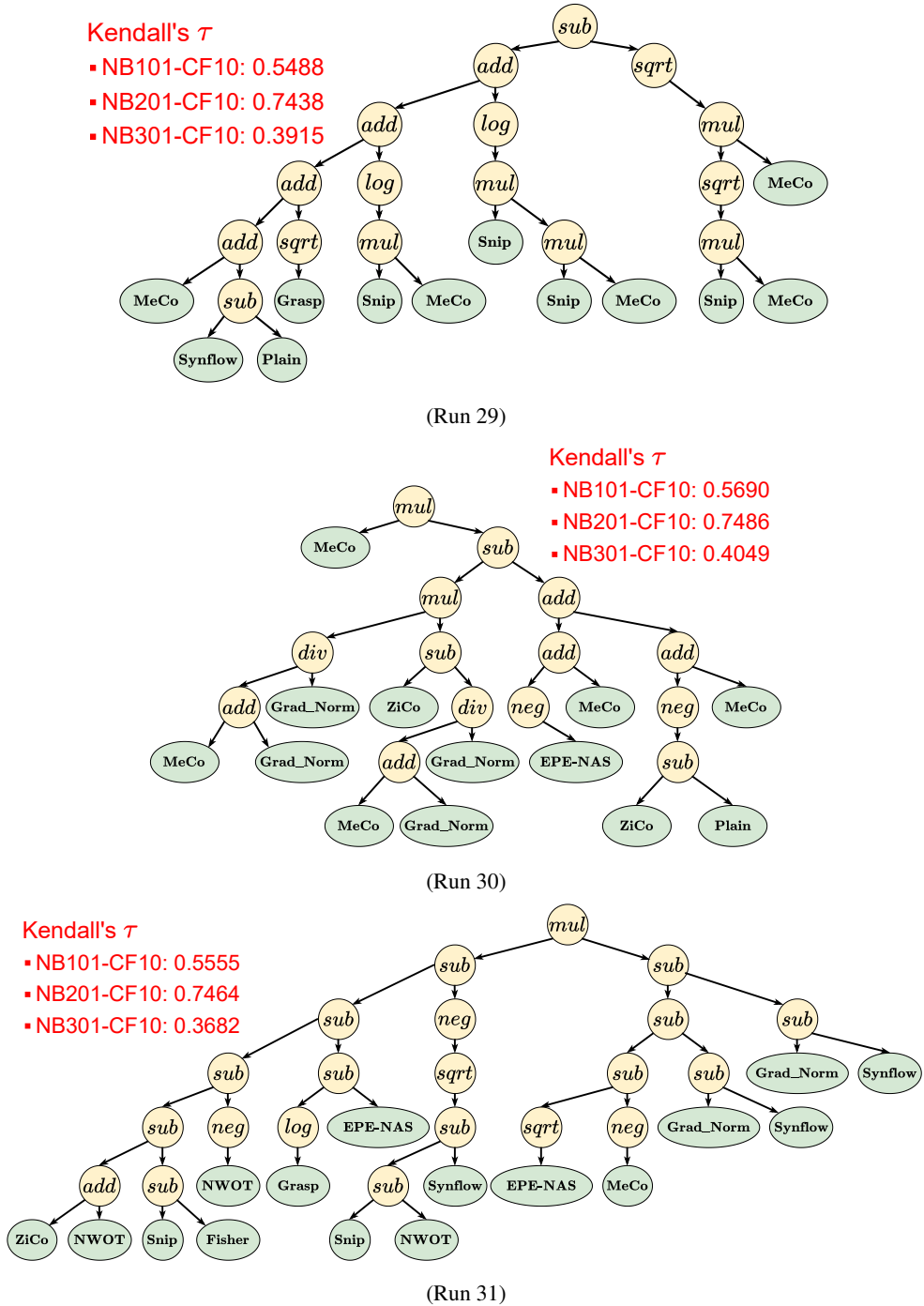

Figure 5: Expression trees of 31 ZC metrics designed by our framework. We also exhibit Kendall's $\tau$ scores across NB101-CF10, NB201-CF10, and NB301-CF10 problems for each ZC metric. The best ZC metric presented in Equation 2 is obtained at the run 12-th.

14

## REFERENCES

Mohamed S. Abdelfattah, Abhinav Mehrotra, Lukasz Dudziak, and Nicholas Donald Lane. Zero-Cost Proxies for Lightweight NAS. In *International Conference on Learning Representations (ICLR)*, 2021.

Tangyu Jiang, Haodi Wang, and Rongfang Bie. MeCo: Zero-Shot NAS with One Data and Single Forward Pass via Minimum Eigenvalue of Correlation. In *Advances in Neural Information Processing Systems (NeurIPS)*, 2023.

Namhoon Lee, Thalaiyasingam Ajanthan, and Philip H. S. Torr. Snip: Single-Shot Network Pruning based on Connection sensitivity. In *International Conference on Learning Representations (ICLR)*, 2019.

Guihong Li, Yuedong Yang, Kartikeya Bhardwaj, and Radu Marculescu. ZiCo: Zero-shot NAS via inverse Coefficient of Variation on Gradients. In *International Conference on Learning Representations (ICLR)*, 2023.

Ming Lin, Pichao Wang, Zhenhong Sun, Hesen Chen, Xiuyu Sun, Qi Qian, Hao Li, and Rong Jin. Zen-NAS: A Zero-Shot NAS for High-Performance Image Recognition. In *International Conference on Computer Vision (ICCV)*, 2021.

Vasco Lopes, Saeid Alirezazadeh, and Luís A. Alexandre. EPE-NAS: Efficient Performance Estimation Without Training for Neural Architecture Search. In *International Conference on Artificial Neural Networks (ICANN)*, 2021.

Joe Mellor, Jack Turner, Amos J. Storkey, and Elliot J. Crowley. Neural Architecture Search without Training. In *International Conference on Machine Learning (ICML)*, 2021.

Joseph Mellor, Jack Turner, Amos J. Storkey, and Elliot J. Crowley. Neural Architecture Search without Training. *CoRR*, 2020. URL https://arxiv.org/abs/2006.04647v1.

Yameng Peng, Andy Song, Haytham M. Fayek, Vic Ciesielski, and Xiaojun Chang. SWAP-NAS: Sample-Wise Activation Patterns for Ultra-fast NAS. In *International Conference on Learning Representations (ICLR)*, 2024.

Hidenori Tanaka, Daniel Kunin, Daniel L. K. Yamins, and Surya Ganguli. Pruning neural networks without any data by iteratively conserving synaptic flow. In *Advances in Neural Information Processing Systems (NeurIPS)*, 2020.

Jack Turner, Elliot J. Crowley, Michael F. P. O'Boyle, Amos J. Storkey, and Gavin Gray. BlockSwap: Fisher-guided Block Substitution for Network Compression on a Budget. In *International Conference on Learning Representations (ICLR)*, 2020.

Chaoqi Wang, Guodong Zhang, and Roger B. Grosse. Picking Winning Tickets Before Training by Preserving Gradient Flow. In *International Conference on Learning Representations (ICLR)*, 2020.



\end{document}